\documentclass[11pt, a4paper, logo, onecolumn, copyright]{googledeepmind}

\pdftrailerid{redacted}

\makeatletter
\renewcommand\bibentry[1]{\nocite{#1}{\frenchspacing\@nameuse{BR@r@#1\@extra@b@citeb}}}
\makeatother

\usepackage{longtable}
\usepackage{dsfont}
\usepackage{gdm-colors}
\usepackage{multirow}
\usepackage{graphicx}
\usepackage{rotating}
\usepackage{geometry}
\usepackage{xcolor}
\usepackage{placeins}
\usepackage{tabularx,tabulary}
\newcolumntype{Y}{>{\centering\arraybackslash}X}
\usepackage{colortbl}
\usepackage{hhline}

\usepackage[numbers, sort&compress, square]{natbib}

\makeatletter
\DeclareRobustCommand\onedot{\futurelet\@let@token\@onedot}
\def\@onedot{\ifx\@let@token.\else.\null\fi}

\def\eg/{\emph{e.g}\onedot} \def\Eg/{\emph{E.g}\onedot}
\def\ie/{\emph{i.e}\onedot} \def\Ie/{\emph{I.e}\onedot}
\def\cf/{\emph{c.f}\onedot} \def\Cf/{\emph{C.f}\onedot}
\def\etc/{\emph{etc}\onedot} \def\vs/{\emph{vs}\onedot}
\def\wrt/{w.r.t\onedot} \def\dof/{d.o.f\onedot}
\def\etal/{\emph{et al}\onedot}

\newcommand{\arxiv}[1]{}
\makeatother
\newcolumntype{L}[1]{>{\raggedright\let\newline\\\arraybackslash\hspace{0pt}}m{#1}} % Left-aligned, fixed-width column

\title{\modelname: A Controllable and Interactive Simulation Framework for Vision Research}
\newcommand{\modelname}{LychSim}

\newcommand{\titlerunning}{LychSim: A Controllable and Interactive Simulation Framework for Vision Research}

\renewcommand{\today}{November 2025}

\author[ \hspace{-0.6ex}]{Wufei Ma}
\author[ \hspace{-0.6ex}]{Chloe Wang}
\author[ \hspace{-0.6ex}]{Siyi Chen}
\author[ \hspace{-0.6ex}]{Jiawei Peng}
\author[ \hspace{-0.6ex}]{Patrick Li}
\author[ \hspace{-0.6ex}]{Alan Yuille}

\affil[ \hspace{-0.6ex}]{Johns Hopkins University}

\usepackage[utf8]{inputenc}
\usepackage{minted}
\usepackage{tcolorbox}
\usepackage{xcolor}
\tcbuselibrary{minted, skins, breakable}

\definecolor{blue-bg}{RGB}{250, 253, 255}    % 极淡的冰蓝色背景
\definecolor{blue-frame}{RGB}{180, 200, 220} % 灰蓝色边框
\definecolor{blue-title}{RGB}{220, 235, 250} % 稍深一点的标题栏背景
\definecolor{blue-text}{RGB}{40, 70, 100}    % 深蓝色标题文字（保证对比度）

\newtcblisting[auto counter]{lightcodefig}[2][]{
  enhanced jigsaw, listing only, listing engine=minted, minted language=#2,
  minted options={fontsize=\small, breaklines, linenos, numbersep=8pt, style=tango, baselinestretch=1.1},
  arc=3pt, boxrule=0.5pt, colback=blue-bg,
  colframe=blue-frame, coltitle=blue-text,
  fonttitle=\bfseries\small\sffamily,
  attach boxed title to top left={yshift=-2mm, xshift=3mm},
  boxed title style={colback=blue-title, colframe=blue-frame, arc=2pt, boxrule=0.5pt},
  left=18pt, #1
}

\newtcblisting[auto counter]{lightcode}[2][]{
  enhanced jigsaw, breakable, listing only, listing engine=minted, minted language=#2,
  minted options={fontsize=\small, breaklines, linenos, numbersep=8pt, style=tango, baselinestretch=1.1},
  arc=3pt, boxrule=0.5pt, colback=blue-bg,
  colframe=blue-frame, coltitle=blue-text,
  fonttitle=\bfseries\small\sffamily,
  attach boxed title to top left={yshift=-2mm, xshift=3mm},
  boxed title style={colback=blue-title, colframe=blue-frame, arc=2pt, boxrule=0.5pt},
  left=18pt, #1
}

\definecolor{orange-bg}{RGB}{255, 253, 248}    % Very light peach
\definecolor{orange-frame}{RGB}{255, 180, 100}  % Vibrant orange border
\definecolor{orange-title}{RGB}{255, 230, 190}  % Light orange title fill
\definecolor{orange-text}{RGB}{230, 81, 0}     % Deep, vibrant orange for text

\newtcblisting[auto counter]{orangecode}[2][]{
  enhanced, breakable, listing only, minted language=#2,
  segmentation style={draw=none}, lower separated=false,
  minted options={fontsize=\small, breaklines, linenos, numbersep=8pt, style=tango, baselinestretch=1.1},
  arc=3pt, boxrule=0.5pt, colback=orange-bg,
  colframe=orange-frame, coltitle=orange-text,
  fonttitle=\bfseries\small\sffamily,
  attach boxed title to top left={yshift=-2mm, xshift=3mm},
  boxed title style={colback=orange-title, colframe=orange-frame, arc=2pt, boxrule=0.5pt},
  left=18pt, #1
}

\begin{abstract}
While self-supervised pretraining has reduced vision systems' reliance on synthetic data, simulation remains an indispensable tool for closed-loop optimization and rigorous out-of-distribution (OOD) evaluation.
However, modern simulation platforms often present steep technical barriers, requiring extensive expertise in computer graphics and game development.
In this work, we present LychSim, a highly controllable and interactive simulation framework built upon Unreal Engine 5 to bridge this gap. LychSim is built around three key designs: (1) a streamlined Python API that abstracts away underlying engine complexities; (2) a procedural data pipeline capable of generating diverse, high-fidelity environments with varying out-of-distribution (OOD) visual challenges, paired with rich 2D and 3D ground truths; and (3) a native integration of the Model Context Protocol (MCP) that transforms the simulator into a dynamic, closed-loop playground for reasoning agentic LLMs.
We further annotate scene-level procedural rules and object-level pose alignments to enable semantically aligned 3D ground truths and automated scene modification.
We demonstrate LychSim's capability across multiple downstream applications, including serving as a synthetic data engine, powering reinforcement learning-based adversarial examiners, and facilitating interactive, language-driven scene layout generation.
To benefit the broader vision community, LychSim will be made publicly available, including full source code and various data annotations.
\end{abstract}

\begin{document}

\maketitle

\section{Introduction} \label{sec:intro}

Recent advancements in self-supervised and weakly-supervised visual pretraining have revolutionized the field of computer vision. Pretraining models~\cite{radford2021learning,he2022masked,zhou2021ibot,caron2021emerging,oquab2023dinov2,simeoni2025dinov3} have demonstrated impressive capabilities to learn rich and transferable visual representations from Internet-scale image and video data, using raw visual contents or naturally occurring text captions.
These advances substantially reduce the amount of labeled data or task-specific fine-tuning required to achieve strong performance across a broad range of downstream tasks, spanning both 2D vision (\textit{e.g.}, classification and segmentation) and 3D vision (\textit{e.g.}, depth estimation, 3D object detection, and pose estimation).
Consequently, the dependence on manually-curated synthetic datasets, which can help mitigate the scarcity of real-world annotations, has diminished as powerful visual representations can now be learned directly from large-scale, unannotated data.

% Why simulation in 2025? (1) analyzing vision systems: comprehensive 2D and 3D ground truths, various ood simulation, analysis of robustness and generalization, (2) interactive simulation for closed-loop training.

Despite the reduced reliance on synthetic data for direct supervised training, the role of simulation remains critically important for computer vision research, as driven by two key objectives. First, simulation environments provide an unparalleled platform for analyzing and understanding complex vision systems. They offer comprehensive and perfectly aligned 2D and 3D ground truths, enable the creation of diverse and controlled Out-of-Distribution (OOD) scenarios, and allow for rigorous analysis of a model's robustness and generalization capabilities in ways that real-world data collection simply cannot replicate.
Second, interactive, high-fidelity simulation is essential for closed-loop training and optimization, especially for embodied AI and robotics. In these applications, agents must learn complex control policies through interaction with their environment, making a realistic and safe virtual playground an indispensable tool for developing and testing advanced, interactive AI systems.

In this work, we present \textcolor{lychred}{LychSim}, a controllable and interactive simulation framework featuring three key designs:
(1) \textbf{Ease of use.} We provide a streamlined Python API that abstracts away various technical complexities in UE5 and C++ development, empowering researchers to script and manipulate high-fidelity 3D scenes without prior computer graphics expertise.
(2) \textbf{A built-in procedural data pipeline with rich 2D and 3D ground truths.} LychSim seamlessly generates diverse environments with various out-of-distribution (OOD) visual challenges, paired with pixel-accurate annotations. Beyond standard labels, our engine models underlying 3D structures and provide ground truths for part segmentation, point maps, and occlusion ratios/relationships for objects extending beyond visible regions.
This unlock new opportunities to explore richer 3D representations and modern 3D learning pipelines.
(3) \textbf{Interactive simulation.} By natively integrating programmatic controls and Model Context Protocol (MCP), LychSim enables algorithms and agentic LLMs to easily navigate, query, and manipulate the 3D world in real-time. This dynamic, closed-loop playground enables many advanced applications, such as RL-based adversarial examiners that systematically identify vision models' weaknesses and interactive, language-driven agentic scene planning.

With the controllable and interactive simulation provided by LychSim, we hope to help advance computer vision research towards a better understanding and more accurate generation of the 3D world.
We believe in the great potential of graphics-based simulation for computer vision research, as a rigorous evaluation framework with diverse 2D and 3D ground truths or a controllable and scalable data engine for model training.
We will release our LychSim publicly, including: (1) \textbf{the complete C++ and Python source code}, and (2) \textbf{associated data annotations}, such as procedural rules for scene generation and pose alignments for object meshes.
%
% Our Python integration abstracts away all the Unreal Engine or C++-related technical complexities, enabling easy use without any prior experiences with computer graphics or Unreal Engine.
%
% We welcome community feedback and contributions, and will remain dedicated to continuously enhancing the simulation with new functionalities to support a wider range of applications.

\begin{figure}[t]
    \centering
    \includegraphics[width=\linewidth]{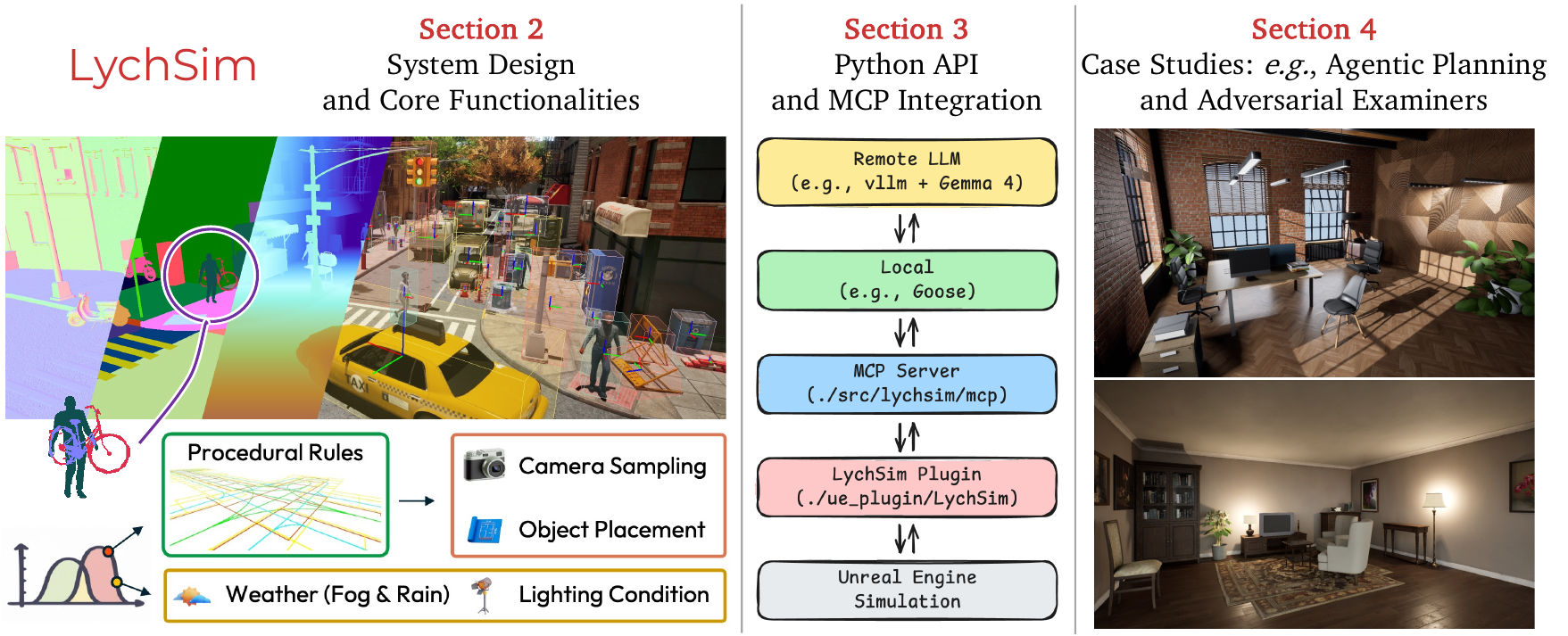}
    \caption{\textbf{We introduce LychSim, a controllable and interactive simulation framework designed for computer vision research.} Our simulation features three key designs: (1) ease of use with a streamlined Python interface, (2) a built-in procedural data pipeline with rich 2D and 3D ground truths, and (3) controllable and interactive simulation.}
    \label{fig:teaser}
\end{figure}

% \paragraph{Code and data release.}
%
% To facilitate reproducibility and community adoption, we publicly release the complete LychSim codebase, including the underlying UE5 \mintinline{}{C++} plugin, the Python scene planning and data generation pipeline, and the Python API with integrated Model Context Protocol (MCP) server support.
%
% Additionally, we provide two specialized data releases to benefit the community: (1) standardized scale and canonical viewpoint alignments for 300 assets across diverse indoor and outdoor environments, and (2) human-annotated procedural rules for 80 scenes, enabling structured and realistic environment generation.

The remainder of this paper is structured as follows.
We introduce the system design and core functionalities of the LychSim simulation system in Section~\ref{sec:design}.
Then we describe the Python API and the Model Context Protocol (MCP) integration in Section~\ref{sec:integration}.
Next in Section~\ref{sec:demo} we present three case compelling studies and demonstrate practical utilities of LychSim in advanced vision research.
Lastly we discuss related works in Section~\ref{sec:related} and summarize our contributions in Section~\ref{sec:conclusions}.

\section{Simulation System Design} \label{sec:design}

\subsection{3D Assets and Data Annotations} \label{sec:assets}

A key advantage of UE5-based simulation systems is direct access to a vast library of high-quality, artist-created 3D assets.
By operating within this native ecosystem, we avoid the rendering artifacts and material inconsistencies that often arise when assets are ported across different simulation platforms.
However, these raw assets are often unstructured and lack a unified representation, making automated manipulation challenging.
To address this and better support advanced computer vision research, we introduce two key data extensions.

\textbf{First, we annotate the category, canonical scale, and pose alignment for the 3D object assets within these scenes.}
These annotations are critical for producing semantically aligned ground-truth 3D object poses and facilitating programmatic object placement and scene manipulation.
\textbf{Second, we define scene-level procedural rules,} such as navigable floor spaces, road areas, pedestrian walks, and dynamic trajectories.
These spatial priors guide the structural generation process, ensuring that newly synthesized layouts remain faithful to the original scene semantics.
The list of 3D assets used in our LychSim and all corresponding data annotations will released publicly.

\subsection{Setting Up 3D Environments}

Setting up realistic and diverse 3D scenes often requires significant human effort, such as creating scene maps, configuring realistic environmental and object lighting, and generating diverse yet plausible 3D object layouts. Prior works explored procedural generation for residential apartments~\cite{procthor,infinigen-indoor}, as well as outdoor environments~\cite{infinigen-natural,yesimworld,deng2024citycraft}.
However, these methods are often constrained to particular domains and object categories, failing to capture the complex, nuanced details of manually curated spaces, such as photorealistic lighting configurations, semantically coherent, physically plausible object layouts, or long-tail diversity and organic randomness of real-world scenes.

\paragraph{A hybrid approach.}
In LychSim, we explore a hybrid approach that incorporates advantages of existing methods.
Specifically, we obtain a variety of 3D scenes from UE5 Fab Asset Marketplace~\cite{fab_marketplace}, encompassing a diverse selection of indoor and outdoor environments that span multiple architectural styles, geographies, and lighting conditions.
This provides us with high-quality, artist-created environments alongside a rich library of object meshes and materials.
%
% We then annotate these scenes with procedural rules defining semantic regions and behaviors, such as floor spaces, road areas, pedestrian walks, and movement trajectories.
%
With the annotated procedural rules and object annotations (see Section~\ref{sec:assets}), our data pipeline subsequently modifies and populates the original environments to generate vast permutations of new scenes.
Finally we also support integration with external 3D scene layouts, such as Infinigen~\cite{infinigen-indoor} and HSSD-200~\cite{khanna2024habitat}, to further enrich our scene diversity.

\paragraph{Levels of visual complexities.}
One advantage of simulation systems is having full control of the 3D scene, producing data with varying levels of visual complexities~\cite{li2023super,ma20253dsrbench}.
With our annotated procedural rules, we further construct targeted sampling pipelines that synthesize challenging, out-of-distribution (OOD) data, featuring uncommon camera viewpoints, severe object occlusions, high-density scenes, and semantically cluttered scenes with objects of the same category densely grouped together.
These out-of-distribution (OOD) data help identify key weaknesses of computer vision models~\cite{ma2026unreal3dspace} and provide valuable fine-tuning data to improve model robustness.

\subsection{Ground Truth Labels}

One advantage of LychSim is its comprehensive collection of 2D and 3D ground-truth annotations, which supports the training and evaluation of a wide range of vision and multi-modal models.
This collection includes standard annotations explored in prior works~\cite{unrealcv,infinigen-indoor}, such as depth maps, instance segmentation, surface normals, point maps, and 2D and 3D object bounding boxes.
In addition, we introduce several novel forms of ground truth that may benefit some emerging areas in computer vision.
We refer the readers to Section~\ref{sec:supp_gt} for qualitative examples of various 2D and 3D ground truths in LychSim.

\paragraph{Beyond visible areas.}
Despite the improved performance and expanded capabilities of modern vision systems, they remain fundamentally limited when dealing with partial occlusion and truncation~\cite{zhao2022ood}.
Addressing this challenge requires moving beyond what is directly observable.
To this end, LychSim explicitly models the underlying 3D scene structure beyond visible regions, enabling fine-grained and quantitative analysis of these failure modes.
Concretely, we capture instance-level depth buffers and perform geometric projection when objects extend outside the image plane. This allows us to accurately estimate per-object occlusion and truncation ratios, as well as recover occlusion relationships between objects. This provides a level of supervision that is difficult to obtain from real-world data.
Figure~\ref{fig:teaser} illustrates the underlying structure of the bicycle that is occluded by the pedestrian.

\paragraph{Part-level segmentation and point maps.} Leveraging the flexibility of the UE5 rendering pipeline, we customize the render targets to directly output object part IDs and per-pixel 3D vertex positions. This enables the extraction of accurate part-level segmentation and dense point maps in a fully automated manner.
The part segmentation maps can be further combined with the visibility information described above to derive fine-grained part-level visibility.
Moreover, the point maps provide precise geometric supervision and align naturally with modern 3D learning pipelines~\cite{wang2024dust3r,wang2025vggt,leroy2024grounding,zhang2024monst3r}.
Together, these annotations open up new opportunities for learning richer object representations that go beyond coarse, instance-level understanding.

\section{Python and Agent Integration} \label{sec:integration}

\subsection{Python Integration}

\begin{figure}[t]
    \centering
    \begin{lightcodefig}[title=LychSim Python API,label={lst:skill}]{python3}
sim = LychSim(server_name="localhost", port=9000)

# Scene layout planning: e.g., from procedural rules
locations, rotations = ..., ...

# Spawn skeletal mesh
sim.add_obj(
    obj_id="reach_truck_01",
    obj_path="/Game/path/to/SKM_Reach_Truck_01a.SKM_Reach_Truck_01a",
    locations["reach_truck_01"], rotations["reach_truck_01"])

# Spawn blueprint vehicle
sim.add_obj(
    obj_id="car_01",
    obj_path="/Game/path/to/BP_vehicle12_Car.BP_vehicle12_Car",
    locations["car_01"], rotations["car_01"],
    adjust_if_possible=True)

# Rendering pixel-level ground truths
img: PIL.Image = sim.get_cam_lit(cam_id=0, warmup=50)
seg: PIL.Image = scene_state.sim.get_cam_seg(cam_id=0)
depth: np.ndarray = sim.get_cam_depth(cam_id=0)
point_map: np.ndarray = sim.get_cam_pointmap(
    cam_id=0, space="opencv")["opencv"]

# Saving object annotations that can fully reconstruct the 3D scene
obj_annots = sim.get_obj_annots()
\end{lightcodefig}
    \caption{\textbf{LychSim Python API code example.} This code snippet demonstrates a unified interface for spawning diverse asset types, such as skeletal meshes and blueprints, by abstracting away underlying engine-level complexities. It demonstrates streamlined rendering of comprehensive 2D and 3D ground truths, including RGB, depth, segmentation, and point maps, alongside the object-level annotations necessary for reconstructing the 3D scene.}
    \label{fig:python}
\end{figure}

Learning to use professional simulation engines like Unreal Engine 5 or Blender presents a significant barrier for many vision researchers, as these tools are often non-intuitive to use and require a substantial investment of time and effort to master.
LychSim addresses this challenge by providing a streamlined Python integration that abstracts away the underlying technical complexities of the engine.
By relieving researchers of the intricacies of computer graphics and $C++$ development, our library enables them to deploy and manipulate simulations without requiring prior experience in game engine architecture.

A particular challenge within the Unreal Engine ecosystem is the varied implementation of 3D assets, which are typically categorized into \mintinline{cpp}{StaticMesh}, \mintinline{cpp}{SkeletalMesh}, or \mintinline{cpp}{Blueprints}.
Standard engine workflows often require distinct procedures to spawn or interact with these different classes, creating friction for automated data generation.
LychSim overcomes this by implementing a unified interface that handles these discrepancies internally. This allows users to utilize the same set of high-level commands to add, edit, or control any object in the scene, regardless of its underlying engine-level representation.

This design philosophy translates into a highly efficient workflow where complex scene manipulations and data generation are reduced to simple Python commands. A researcher can programmatically spawn assets, adjust their 3D coordinates, or remove them from the environment using a straightforward and consistent API. Crucially, LychSim enables the generation of comprehensive ground truths with minimal effort; with simple function calls, the system renders and retrieves synchronized RGB images, depth maps, instance-level segmentations, and point maps. This ensures that the simulation serves as a robust and accessible data engine that is both controllable and easy to iterate upon for various vision tasks.
%
% We showcase an example simulation workflow in Figure~\ref{fig:python} and refer the readers to the \href{https://wufeim.github.io/LychSim/docs/index.html}{LychSim documentation page} for a full catalog of the library's capabilities and supported functions.
In Figure~\ref{fig:python} we showcase an example LychSim simulation workflow using only the Python interface.

\subsection{Model Context Protocol (MCP)}

Recent advancements in Large Language Models (LLMs), such as Claude Opus 4.6~\cite{opus_4_6} and Gemma 4~\cite{gemma_4_31b_2026}, have shifted the focus toward systems that can autonomously use tools to solve complex tasks. Integrating LychSim with Model Context Protocol (MCP) is an essential step to bridge the gap between reasoning agentic LLMs and the 3D simulation environment.
With a standardized interface, we enable agents to move beyond static data processing and engage in ``closed-loop'' interactions with the 3D world.
%
% We find that this integration transforms the simulation into a truly controllable platform where vision systems can be analyzed through dynamic, natural language-driven scenarios.

We implement the MCP integration by hosting a dedicated server that exposes our Python API as a suite of standardized agentic tools. We provide a comprehensive toolset that allows an AI agent to navigate within the scene, query structured scene state, capture real-time visual renderings, and manipulate objects programmatically. We refer the readers to Section~\ref{sec:supp_mcp} with more technical details on the MCP design.
%
%Notably, this visual loop allows the agent to take screenshots of the rendered environment and self-correct any placement errors or overlaps.
%
% We define specific "skills" that guide the agent through a multi-step workflow: parsing a scene specification, planning functional zones, placing anchor objects, and verifying the layout through visual feedback. We emphasize the importance of this visual loop, as it allows the agent to take screenshots of the rendered environment and self-correct any placement errors or overlaps.
%
In Section~\ref{sec:demo}, we demonstrate that the LychSim MCP integration enables a wide range of interactive applications, from adversarial examiners (Section~\ref{sec:exp_adversarial}) to interactive scene layout planning and generation (Section~\ref{sec:exp_interactive}).

\begin{figure}
    \centering
    \includegraphics[width=\linewidth]{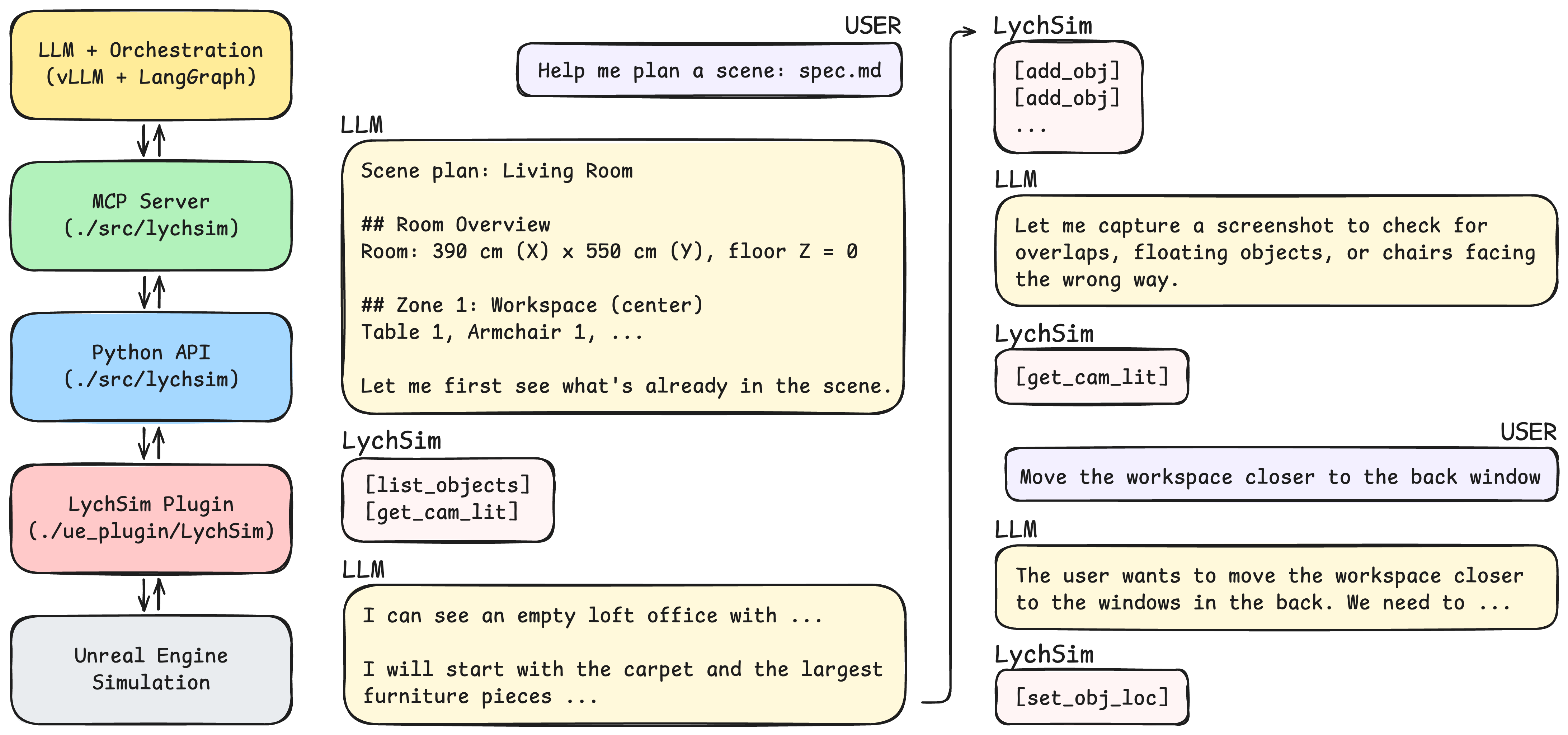}
    \caption{\textbf{Agentic integration and interactive scene planning.} \textbf{Left}: Our LychSim provides Python and MCP server that allows seemless integration with other}
    \label{fig:interactive_flow}
\end{figure}

\section{Case Studies} \label{sec:demo}

\begin{figure}[t]
    \centering
    \begin{minipage}{0.38\linewidth}
    \begin{lightcodefig}[title=Adversarial Examiner for SAM,label={lst:adversarial}]{python3}
sim = LychSim(...)
p = GaussianPolicy()
for _ in range(n_steps):
    pose = p.generate()
    sim.set_pose(pose)
    img, mask = \
      sim.render(...)
    rwd = rwd_func(
        sam(img), mask)
    p.update(pose, rwd)
\end{lightcodefig}
    \end{minipage}
    \hfill
    \begin{minipage}{0.6\linewidth}
    \includegraphics[width=\textwidth]{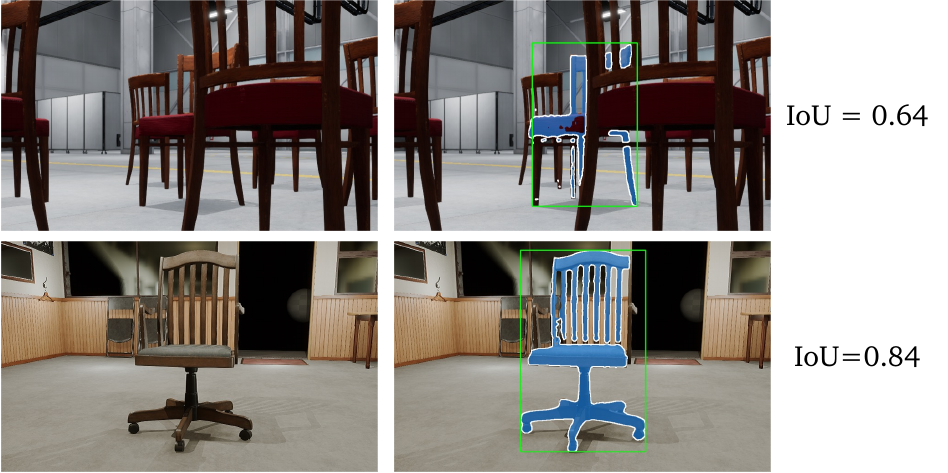}
    \end{minipage}
    \caption{\textbf{Case study of adversarial examiner for instance segmentation.} \textbf{Left}: pseudo-code for RL-based adversarial examiner~\cite{shu2020identifying,liu2023poseexaminer} running with LychSim simulation. \textbf{Right}:}
    \label{fig:adv_exam}
\end{figure}

\subsection{LychSim as Synthetic Data Engine} \label{sec:exp_data_engine}

LychSim introduces a controllable and procedural simulation pipeline that enables the generation of high-fidelity synthetic data with comprehensive 2D and 3D ground truths.
We highlight two practical applications of this data: (1) diagnosing the weaknesses of current spatial vision-language models (VLMs), and (2) serving as a scalable data engine for VLM post-training.

\paragraph{For evaluation and analysis.}
Despite the domain gap between synthetic and real data, synthetic benchmarks have been widely adopted in vision research. They offer unparalleled controllability with varying visual complexities~\cite{johnson2017clevr,girdhar2019cater,li2023super,wang20233d}, abundant pixel-accurate 3D ground truths~\cite{ros2016synthia,dosovitskiy2017carla,roberts2021hypersim}, and even interactive 3D environments for embodied AI research~\cite{ai2thor,puig2023habitat3}.
Some more recent works built on LychSim and studied more fine-grained and challenging problems in multi-modal reasoning. Unreal3DSpace~\cite{ma2026unreal3dspace} analyzed failure patterns in spatial reasoning through models' chain-of-thought trajectory. PerceptualTaxonomy~\cite{lee2025perceptual} required the model to infer task-relevant properties from 3D scenes and enable goal-directed reasoning.

\paragraph{For model training.}
LychSim can also serve as a highly scalable synthetic data framework for generating post-training data that enhance various 2D and 3D spatial understanding abilities of vision-language models.
Prior successes in this area, including SAT~\cite{ray2024sat}, ScanForgeQA~\cite{zhang2025spatial}, and SIMS-V~\cite{brown2025sims}, demonstrate that scalable, high-fidelity simulation can be effectively integrated into the post-training loop and substantially improve spatial understanding performance.

\subsection{Adversarial Examiners} \label{sec:exp_adversarial}

Standard datasets are often limited to a narrow subset of the broader real-world parameter space. This restriction introduces bias in evaluation, such as in terms of object appearance and shape~\cite{zhao2022ood} or object 3D pose~\cite{ma2024generating}.
Adversarial examiners~\cite{shu2020identifying} address this limitation by systematically exploring the parameter space in simulation and revealing the weaknesses in vision models.

Following prior works~\cite{shu2020identifying,Ruiz_2022_CVPR,liu2023poseexaminer}, 
we adopt a reinforcement learning (RL)-based adversarial examiner and train a Gaussian policy to identify the weaknesses of Segment Anything~\cite{kirillov2023segment}.
Specifically, the adversarial examiner explores different 3D camera viewpoints within a sphere around the target object and is optimized to minimize the intersection-over-union (IoU) of SAM predictions.
Failure examples in Figure~\ref{fig:adv_exam} demonstrate that adversarial examiner can effectively capture model weaknesses even on common objects in simple environments.

\subsection{Interactive Scene Planning and Generation} \label{sec:exp_interactive}

With the improved spatial awareness of vision-language models (VLMs)~\cite{chen2024spatialvlm,cheng2024spatialrgpt,ma2025spatialllm,ma2026spatialreasoner}, we have seen great progress in 3D scene layout generation from natural language~\cite{feng2023layoutgpt,yang2024holodeck,ccelen2024design,sun2025layoutvlm}. The models are capable of generating realistic and physically-viable 3D layouts following the descriptions in the prompt. Beyond these feed-forward models, we demonstrate an example of interactive scene planning and generation using Opus 4.6~\cite{opus_4_6} and Gemma 4~\cite{gemma_4_31b_2026}.

\begin{figure}[t]
    \centering
    \includegraphics[width=\linewidth]{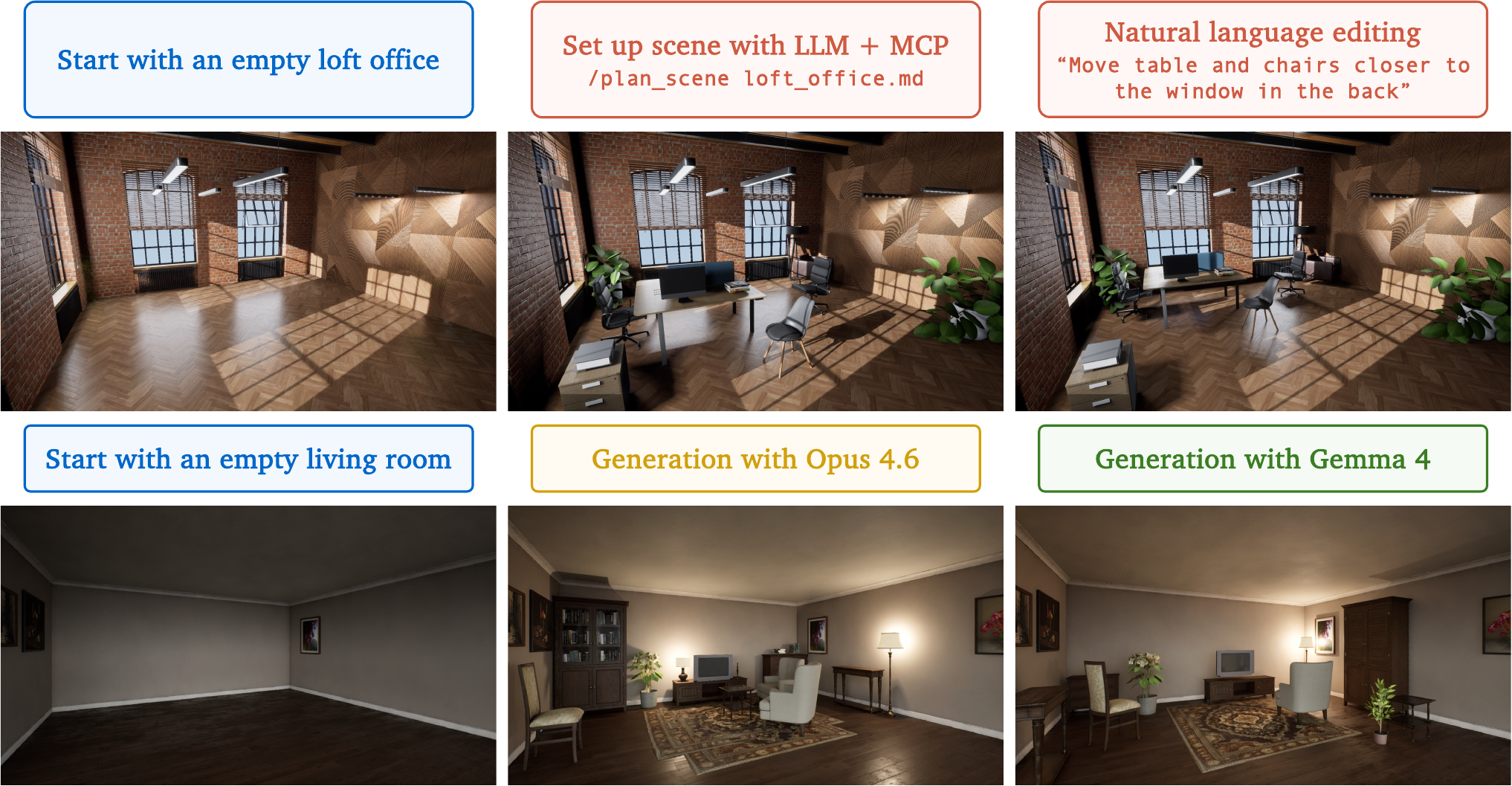}
    \caption{\textbf{Interactive scene planning.} Top: interactive scene planning with agentic skill and MCP integration, as well as natural language control. Bottom: Scene layout generation with Claude Opus 4.6~\cite{opus_4_6} and Gemma 4~\cite{gemma_4_31b_2026}. See agent skill definition \mintinline{text}{SKILL.md} in Code~\ref{lst:plan_room} and scene planning specifications \mintinline{text}{loft_office.md} in Code~\ref{lst:office}.}
    \label{fig:plan_scene}
\end{figure}

As illustrated in Figure~\ref{fig:interactive_flow}, our interactive environment is built on the Unreal Engine 5 and the LychSim plugin, interfaces with the agentic LLM through an MCP server.
The model is provided with a scene specification file that captures user requirements (see Code~\ref{lst:office}), together with a skill file containing lightweight guidance and a list of available MCP tools (see Code~\ref{lst:plan_room}).
From the results in Figure~\ref{fig:interactive_flow} and Figure~\ref{fig:plan_scene}, we demonstrate that the agentic model can (1) plan a complete scene that follows the requirements in the specification file, (2) navigate and inspect the scene from multiple camera viewpoints to identify and correct physically implausible layouts, such as a vase floating in midair, and (3) edit the generated scene following user requests in a multi-turn conversation.

We also note several failure patterns in this pipeline, including physically implausible layouts and object collisions, which largely attribute to the limited spatial reasoning capabilities of current state-of-the-art models. Nevertheless, we believe this is a promising research direction for interactive 3D scene design.

\section{Related Works} \label{sec:related}

\paragraph{Synthetic data engine.}

The role of simulation remains critically important for computer vision research, specifically for providing perfectly aligned ground truths and closed-loop environments for training and optimization. Existing synthetic data frameworks can be broadly categorized by their underlying platforms:
(1) \textit{Blender-based}: Recent works like Infinigen~\cite{infinigen-indoor,infinigen-natural} offer the ability to generate infinite photorealistic worlds through procedural generation.
InfiniBench~\cite{wang2025infinibench} extended Infinigen with Gemini-2.5-Pro for scene constraint generation, producing scenes with varying scene complexities.
Another line of works explored Blender as the visual interface for LLM-based reconstruction and generation~\cite{kulits2024rethinking,yin2026visionasinversegraphicsagentinterleavedmultimodal,hu2024scenecraft}.
(2) \textit{Unity-based}:
(3) \textit{Unreal Engine-based}: Early works explored UE as the data engine and built synthetic datasets by utilizing high quality 3D assets and modern rendering engine~\cite{virtualkitti,kim2019synthesizing,Tosi2021CVPR}. UnrealCV~\cite{unrealcv} developed a plugin in Unreal Engine that enables communication between Python clients and UE backend.
More recent works, such as UnrealZoo~\cite{zhong2025unrealzoo} and SimWorld~\cite{yesimworld} extended this framework with rich human motions and city 3D scenes, evaluating various embodied AI algorithms on real-world tasks.
Our LychSim built on UnrealCV and largely extended it with more 2D and 3D ground truths, procedural scene generation, and integration with agentic LLMs.
(4) \textit{Physics-focused}: Another line of works builds on platforms such as NVIDIA Isaac Sim, PyBullet~\cite{coumans2021}, and MuJoCo~\cite{todorov2012mujoco}, which are heavily optimized for rigid-body dynamics, contact physics, and high-frequency control loops.
While their accurate and reproducible physics engines make them highly suitable for robotics and reinforcement learning, they often lack the visual diversity and high-fidelity rendering found in parallel, vision-centric simulation efforts.
% With accurate and reproducible physics simulation, these simulation systems are suitable for a range of robitics reinforcement learning, yet they often lack visual diversity and fidelity as in parallel efforts.

% \paragraph{Simulation for robotics and reinforcement learning.} Alongside visual data engines, simulation plays a pivotal role in robotics and reinforcement learning (RL). Platforms such as NVIDIA Isaac Sim, PyBullet~\cite{coumans2021}, and MuJoCo~\cite{todorov2012mujoco} are heavily optimized for rigid-body dynamics, contact physics, and high-frequency control loops. While there is a growing intersection between visual perception and robotics—often studied under the umbrella of embodied AI—the primary objectives of these robotic simulators differ fundamentally from environments designed purely for vision training.

% While engines like Isaac Sim do offer advanced ray-tracing capabilities, authoring diverse, highly realistic scenes purely for vision pre-training or out-of-distribution (OOD) evaluation remains resource-intensive. LychSim is explicitly tailored to bridge this gap for the vision community. By prioritizing high-fidelity Unreal Engine 5 rendering, procedural scene diversity, and extensive 2D/3D semantic annotations, LychSim decouples the complexity of deep visual understanding from the rigid constraints of high-fidelity physics simulation, providing an uncompromised data engine for vision systems.

\paragraph{Analyzing vision systems in simulation.}
While real-world datasets are indispensable for training and evaluation, they are often limited in dataset scale, annotation quality, and lack of fine-grained control that often obscure the underlying failure modes of vision systems~\cite{ma2024imagenet3d,zhao2022ood,zhao2024ood,ma20253dsrbench}.
Simulation addresses this limitation by providing perfect ground-truth annotations and the ability to systematically isolate specific visual factors, such as visual complexities~\cite{li2023super,wang2025infinibench,ma2026unreal3dspace}, occlusion~\cite{ma2026unreal3dspace}, and camera viewpoints~\cite{ma20253dsrbench}.
Early diagnostic datasets~\cite{johnson2017clevr,yi2019clevrer,girdhar2019cater,li2023super} pioneered this approach, utilizing generated scenes to rigorously isolate and evaluate compositional visual reasoning.
More recent efforts have scaled synthetic evaluation to richer, more dynamic environments, such as evaluating complex spatial comprehension across a spectrum of scene and problem difficulties~\cite{ray2024sat,wang2025spatial457,wang2025infinibench}.

\paragraph{Automated scene layout generation.}
The automated synthesis of plausible 3D environments has traditionally been explored through two primary paradigms: (1) \textit{rule-based procedural generation}:
Frameworks such as ProcTHOR~\cite{procthor}, Infinigen~\cite{infinigen-indoor,infinigen-natural}, and InfiniBench~\cite{wang2025infinibench} utilize sophisticated programmatic constraints and spatial algorithms to synthesize vast quantities of structurally viable indoor and outdoor scenes. While highly scalable, these methods are constrained by the quality and quantity of the crafted rules, which may not adapt well to highly specific instruction-based intents.
(2) \textit{LLM-based agentic methods}: Leveraging the deep semantic and spatial reasoning capabilities of foundation models, approaches including SceneCraft~\cite{hu2024scenecraft}, LayoutGPT~\cite{feng2023layoutgpt}, Holodeck~\cite{yang2024holodeck}, I-design~\cite{ccelen2024design}, LayoutVLM~\cite{sun2025layoutvlm}, and CityCraft~\cite{deng2024citycraft} translate natural language prompts directly into 3D spatial layouts.
LychSim serves as a unifying framework for both paradigms: we have built-in rule-based procedural generation pipelines, while the native MCP integration provides an ideal closed-loop environment for executing and evaluating multi-turn, LLM-driven spatial planning.

\section{Conclusions} \label{sec:conclusions}

In this work, we introduce LychSim, a highly controllable and interactive simulation framework designed to bridge the gap between complex 3D graphics engines and the evolving needs of computer vision research.
By combining a streamlined Python API, a robust procedural data pipeline, and native Model Context Protocol (MCP) integration, LychSim provides an accessible, powerful, and closed-loop environment.
Moving beyond standard annotations, our system explicitly models underlying 3D structures to deliver novel ground truths—such as dense point maps, part-level segmentations, and precise occlusion metrics—unlocking new avenues for robust 3D representation learning.
Through three case studies, we demonstrate its versatility in synthesizing challenging out-of-distribution scenarios, empowering RL-based adversarial examiners to uncover model vulnerabilities, and facilitating fully interactive, language-driven scene planning with modern agentic LLMs.
To benefit the broader vision community, we will publicly release the complete LychSim framework, including the C++ and Python source code as well as our procedural rules and object-level annotations.

\section*{Acknowledgements} \label{sec:acknowledge}

LychSim is built on the architecture of UnrealCV~\cite{qiu2017unrealcv}, which exposes Unreal Engine to external Python clients. LychSim extends the plugin into a full interactive simulation framework with new functionalities, procedural generation, and native Python/MCP integration for agentic research.

WM and AY acknowledge support from ONR with N00014-23-1-2641 National Eye Institute (NEI) with Award ID: R01EY037193. This work was also supported in part by the Whiting School of Engineering at Johns Hopkins University.

\FloatBarrier

% Bibliography components
\bibliographystyle{abbrvnat}
\nobibliography*
\bibliography{document}

\begin{thebibliography}{66}
\providecommand{\natexlab}[1]{#1}
\providecommand{\url}[1]{\texttt{#1}}
\expandafter\ifx\csname urlstyle\endcsname\relax
  \providecommand{\doi}[1]{doi: #1}\else
  \providecommand{\doi}{doi: \begingroup \urlstyle{rm}\Url}\fi

\bibitem[{Anthropic}(2026)]{opus_4_6}
{Anthropic}.
\newblock Claude {O}pus 4.6, 2026.
\newblock URL \url{https://www.anthropic.com/news/claude-opus-4-6}.
\newblock Accessed: 2026-04-11.

\bibitem[Brown et~al.(2025)Brown, Ray, Krishna, Girshick, Fergus, and
  Xie]{brown2025sims}
E.~Brown, A.~Ray, R.~Krishna, R.~Girshick, R.~Fergus, and S.~Xie.
\newblock Sims-v: Simulated instruction-tuning for spatial video understanding.
\newblock \emph{arXiv preprint arXiv:2511.04668}, 2025.

\bibitem[Caron et~al.(2021)Caron, Touvron, Misra, J{\'e}gou, Mairal,
  Bojanowski, and Joulin]{caron2021emerging}
M.~Caron, H.~Touvron, I.~Misra, H.~J{\'e}gou, J.~Mairal, P.~Bojanowski, and
  A.~Joulin.
\newblock Emerging properties in self-supervised vision transformers.
\newblock In \emph{Proceedings of the IEEE/CVF international conference on
  computer vision}, pages 9650--9660, 2021.

\bibitem[{\c{C}}elen et~al.(2024){\c{C}}elen, Han, Schindler, Van~Gool, Armeni,
  Obukhov, and Wang]{ccelen2024design}
A.~{\c{C}}elen, G.~Han, K.~Schindler, L.~Van~Gool, I.~Armeni, A.~Obukhov, and
  X.~Wang.
\newblock I-design: Personalized llm interior designer.
\newblock In \emph{European Conference on Computer Vision}, pages 217--234.
  Springer, 2024.

\bibitem[Chen et~al.(2024)Chen, Xu, Kirmani, Ichter, Sadigh, Guibas, and
  Xia]{chen2024spatialvlm}
B.~Chen, Z.~Xu, S.~Kirmani, B.~Ichter, D.~Sadigh, L.~Guibas, and F.~Xia.
\newblock Spatialvlm: Endowing vision-language models with spatial reasoning
  capabilities.
\newblock In \emph{Proceedings of the IEEE/CVF Conference on Computer Vision
  and Pattern Recognition}, pages 14455--14465, 2024.

\bibitem[Cheng et~al.(2024)Cheng, Yin, Fu, Guo, Yang, Kautz, Wang, and
  Liu]{cheng2024spatialrgpt}
A.-C. Cheng, H.~Yin, Y.~Fu, Q.~Guo, R.~Yang, J.~Kautz, X.~Wang, and S.~Liu.
\newblock Spatialrgpt: Grounded spatial reasoning in vision-language models.
\newblock \emph{Advances in Neural Information Processing Systems},
  37:\penalty0 135062--135093, 2024.

\bibitem[Coumans and Bai(2016--2021)]{coumans2021}
E.~Coumans and Y.~Bai.
\newblock Pybullet, a python module for physics simulation for games, robotics
  and machine learning.
\newblock \url{http://pybullet.org}, 2016--2021.

\bibitem[Deitke et~al.(2022)Deitke, VanderBilt, Herrasti, Weihs, Salvador,
  Ehsani, Han, Kolve, Farhadi, Kembhavi, and Mottaghi]{procthor}
M.~Deitke, E.~VanderBilt, A.~Herrasti, L.~Weihs, J.~Salvador, K.~Ehsani,
  W.~Han, E.~Kolve, A.~Farhadi, A.~Kembhavi, and R.~Mottaghi.
\newblock {ProcTHOR: Large-Scale Embodied AI Using Procedural Generation}.
\newblock In \emph{NeurIPS}, 2022.
\newblock Outstanding Paper Award.

\bibitem[Deng et~al.(2024)Deng, Chai, Huang, Zhao, Huang, Gao, Guo, Hao, Hu,
  Hwang, et~al.]{deng2024citycraft}
J.~Deng, W.~Chai, J.~Huang, Z.~Zhao, Q.~Huang, M.~Gao, J.~Guo, S.~Hao, W.~Hu,
  J.-N. Hwang, et~al.
\newblock Citycraft: A real crafter for 3d city generation.
\newblock \emph{arXiv preprint arXiv:2406.04983}, 2024.

\bibitem[Dosovitskiy et~al.(2017)Dosovitskiy, Ros, Codevilla, Lopez, and
  Koltun]{dosovitskiy2017carla}
A.~Dosovitskiy, G.~Ros, F.~Codevilla, A.~Lopez, and V.~Koltun.
\newblock Carla: An open urban driving simulator.
\newblock In \emph{Conference on robot learning}, pages 1--16. PMLR, 2017.

\bibitem[{Epic Games}(2026)]{fab_marketplace}
{Epic Games}.
\newblock Fab asset marketplace, 2026.
\newblock URL \url{https://www.fab.com/}.
\newblock Unified marketplace for digital assets.

\bibitem[Feng et~al.(2023)Feng, Zhu, Fu, Jampani, Akula, He, Basu, Wang, and
  Wang]{feng2023layoutgpt}
W.~Feng, W.~Zhu, T.-j. Fu, V.~Jampani, A.~Akula, X.~He, S.~Basu, X.~E. Wang,
  and W.~Y. Wang.
\newblock Layoutgpt: Compositional visual planning and generation with large
  language models.
\newblock \emph{Advances in Neural Information Processing Systems},
  36:\penalty0 18225--18250, 2023.

\bibitem[Gaidon et~al.(2016)Gaidon, Wang, Cabon, and Vig]{virtualkitti}
A.~Gaidon, Q.~Wang, Y.~Cabon, and E.~Vig.
\newblock Virtual worlds as proxy for multi-object tracking analysis.
\newblock In \emph{CVPR}, 2016.

\bibitem[Girdhar and Ramanan(2019)]{girdhar2019cater}
R.~Girdhar and D.~Ramanan.
\newblock Cater: A diagnostic dataset for compositional actions and temporal
  reasoning.
\newblock \emph{arXiv preprint arXiv:1910.04744}, 2019.

\bibitem[{Google DeepMind}(2026)]{gemma_4_31b_2026}
{Google DeepMind}.
\newblock Gemma-4-31b, 2026.
\newblock URL \url{https://huggingface.co/google/gemma-4-31B}.
\newblock Hugging Face Model Card.

\bibitem[He et~al.(2022)He, Chen, Xie, Li, Doll{\'a}r, and
  Girshick]{he2022masked}
K.~He, X.~Chen, S.~Xie, Y.~Li, P.~Doll{\'a}r, and R.~Girshick.
\newblock Masked autoencoders are scalable vision learners.
\newblock In \emph{Proceedings of the IEEE/CVF conference on computer vision
  and pattern recognition}, pages 16000--16009, 2022.

\bibitem[Hu et~al.(2024)Hu, Iscen, Jain, Kipf, Yue, Ross, Schmid, and
  Fathi]{hu2024scenecraft}
Z.~Hu, A.~Iscen, A.~Jain, T.~Kipf, Y.~Yue, D.~A. Ross, C.~Schmid, and A.~Fathi.
\newblock Scenecraft: An llm agent for synthesizing 3d scenes as blender code.
\newblock In \emph{Forty-first International Conference on Machine Learning},
  2024.

\bibitem[Johnson et~al.(2017)Johnson, Hariharan, Van Der~Maaten, Fei-Fei,
  Lawrence~Zitnick, and Girshick]{johnson2017clevr}
J.~Johnson, B.~Hariharan, L.~Van Der~Maaten, L.~Fei-Fei, C.~Lawrence~Zitnick,
  and R.~Girshick.
\newblock Clevr: A diagnostic dataset for compositional language and elementary
  visual reasoning.
\newblock In \emph{Proceedings of the IEEE conference on computer vision and
  pattern recognition}, pages 2901--2910, 2017.

\bibitem[Khanna et~al.(2024)Khanna, Mao, Jiang, Haresh, Shacklett, Batra,
  Clegg, Undersander, Chang, and Savva]{khanna2024habitat}
M.~Khanna, Y.~Mao, H.~Jiang, S.~Haresh, B.~Shacklett, D.~Batra, A.~Clegg,
  E.~Undersander, A.~X. Chang, and M.~Savva.
\newblock Habitat synthetic scenes dataset (hssd-200): An analysis of 3d scene
  scale and realism tradeoffs for objectgoal navigation.
\newblock In \emph{Proceedings of the IEEE/CVF Conference on Computer Vision
  and Pattern Recognition}, pages 16384--16393, 2024.

\bibitem[Kim et~al.(2019)Kim, Peven, Qiu, Yuille, and
  Hager]{kim2019synthesizing}
T.~S. Kim, M.~Peven, W.~Qiu, A.~Yuille, and G.~D. Hager.
\newblock Synthesizing attributes with unreal engine for fine-grained activity
  analysis.
\newblock In \emph{2019 IEEE Winter Applications of Computer Vision Workshops
  (WACVW)}, pages 35--37. IEEE, 2019.

\bibitem[Kirillov et~al.(2023)Kirillov, Mintun, Ravi, Mao, Rolland, Gustafson,
  Xiao, Whitehead, Berg, Lo, et~al.]{kirillov2023segment}
A.~Kirillov, E.~Mintun, N.~Ravi, H.~Mao, C.~Rolland, L.~Gustafson, T.~Xiao,
  S.~Whitehead, A.~C. Berg, W.-Y. Lo, et~al.
\newblock Segment anything.
\newblock In \emph{Proceedings of the IEEE/CVF international conference on
  computer vision}, pages 4015--4026, 2023.

\bibitem[Kolve et~al.(2017)Kolve, Mottaghi, Han, VanderBilt, Weihs, Herrasti,
  Gordon, Zhu, Gupta, and Farhadi]{ai2thor}
E.~Kolve, R.~Mottaghi, W.~Han, E.~VanderBilt, L.~Weihs, A.~Herrasti, D.~Gordon,
  Y.~Zhu, A.~Gupta, and A.~Farhadi.
\newblock {AI2-THOR: An Interactive 3D Environment for Visual AI}.
\newblock \emph{arXiv}, 2017.

\bibitem[Kulits et~al.(2024)Kulits, Feng, Liu, Abrevaya, and
  Black]{kulits2024rethinking}
P.~Kulits, H.~Feng, W.~Liu, V.~F. Abrevaya, and M.~J. Black.
\newblock Re-thinking inverse graphics with large language models.
\newblock \emph{Transactions on Machine Learning Research}, 2024.
\newblock ISSN 2835-8856.
\newblock URL \url{https://openreview.net/forum?id=u0eiu1MTS7}.

\bibitem[Lee et~al.(2025)Lee, Wang, Peng, Ye, Zheng, Zhang, Wang, Ma, Chen,
  Chou, et~al.]{lee2025perceptual}
J.~Lee, X.~Wang, J.~Peng, L.~Ye, Z.~Zheng, T.~Zhang, T.~Wang, W.~Ma, S.~Chen,
  Y.-C. Chou, et~al.
\newblock Perceptual taxonomy: Evaluating and guiding hierarchical scene
  reasoning in vision-language models.
\newblock \emph{arXiv preprint arXiv:2511.19526}, 2025.

\bibitem[Leroy et~al.(2024)Leroy, Cabon, and Revaud]{leroy2024grounding}
V.~Leroy, Y.~Cabon, and J.~Revaud.
\newblock Grounding image matching in 3d with mast3r.
\newblock In \emph{European conference on computer vision}, pages 71--91.
  Springer, 2024.

\bibitem[Li et~al.(2023)Li, Wang, Stengel-Eskin, Kortylewski, Ma, Van~Durme,
  and Yuille]{li2023super}
Z.~Li, X.~Wang, E.~Stengel-Eskin, A.~Kortylewski, W.~Ma, B.~Van~Durme, and
  A.~L. Yuille.
\newblock Super-clevr: A virtual benchmark to diagnose domain robustness in
  visual reasoning.
\newblock In \emph{Proceedings of the IEEE/CVF conference on computer vision
  and pattern recognition}, pages 14963--14973, 2023.

\bibitem[Liu et~al.(2023)Liu, Kortylewski, and Yuille]{liu2023poseexaminer}
Q.~Liu, A.~Kortylewski, and A.~L. Yuille.
\newblock Poseexaminer: Automated testing of out-of-distribution robustness in
  human pose and shape estimation.
\newblock In \emph{Proceedings of the IEEE/CVF Conference on Computer Vision
  and Pattern Recognition}, pages 672--681, 2023.

\bibitem[Ma et~al.(2024{\natexlab{a}})Ma, Liu, Wang, Wang, Yuan, Zhang, Xiao,
  Zhang, Lu, Duan, Qi, Kortylewski, Liu, and Yuille]{ma2024generating}
W.~Ma, Q.~Liu, J.~Wang, A.~Wang, X.~Yuan, Y.~Zhang, Z.~Xiao, G.~Zhang, B.~Lu,
  R.~Duan, Y.~Qi, A.~Kortylewski, Y.~Liu, and A.~Yuille.
\newblock Generating images with 3d annotations using diffusion models.
\newblock In \emph{The Twelfth International Conference on Learning
  Representations}, 2024{\natexlab{a}}.
\newblock URL \url{https://openreview.net/forum?id=XlkN11Xj6J}.

\bibitem[Ma et~al.(2024{\natexlab{b}})Ma, Zhang, Liu, Zeng, Kortylewski, Liu,
  and Yuille]{ma2024imagenet3d}
W.~Ma, G.~Zhang, Q.~Liu, G.~Zeng, A.~Kortylewski, Y.~Liu, and A.~Yuille.
\newblock Imagenet3d: Towards general-purpose object-level 3d understanding.
\newblock \emph{Advances in Neural Information Processing Systems},
  37:\penalty0 96127--96149, 2024{\natexlab{b}}.

\bibitem[Ma et~al.(2025{\natexlab{a}})Ma, Chen, Zhang, Chou, Chen, de~Melo, and
  Yuille]{ma20253dsrbench}
W.~Ma, H.~Chen, G.~Zhang, Y.-C. Chou, J.~Chen, C.~de~Melo, and A.~Yuille.
\newblock 3dsrbench: A comprehensive 3d spatial reasoning benchmark.
\newblock In \emph{Proceedings of the IEEE/CVF International Conference on
  Computer Vision}, pages 6924--6934, 2025{\natexlab{a}}.

\bibitem[Ma et~al.(2025{\natexlab{b}})Ma, Ye, de~Melo, Yuille, and
  Chen]{ma2025spatialllm}
W.~Ma, L.~Ye, C.~M. de~Melo, A.~Yuille, and J.~Chen.
\newblock Spatialllm: A compound 3d-informed design towards
  spatially-intelligent large multimodal models.
\newblock In \emph{Proceedings of the Computer Vision and Pattern Recognition
  Conference}, pages 17249--17260, 2025{\natexlab{b}}.

\bibitem[Ma et~al.(2026{\natexlab{a}})Ma, Cen, Shen, Lee, Begiristain, Zhuang,
  Peng, Yu, Song, Qi, Shu, Kortylewski, and Yuille]{ma2026unreal3dspace}
W.~Ma, S.~Cen, J.~Shen, R.~Lee, L.~Begiristain, Y.~Zhuang, J.~Peng, Z.~Yu,
  T.~Song, X.~Qi, T.~Shu, A.~Kortylewski, and A.~Yuille.
\newblock Unrealspace: Analyzing spatial understanding and reasoning in
  controllable simulation.
\newblock In \emph{Findings of the Computer Vision and Pattern Recognition
  Conference}, 2026{\natexlab{a}}.

\bibitem[Ma et~al.(2026{\natexlab{b}})Ma, Chou, Liu, Wang, de~Melo, Xie, and
  Yuille]{ma2026spatialreasoner}
W.~Ma, Y.-C. Chou, Q.~Liu, X.~Wang, C.~de~Melo, J.~Xie, and A.~Yuille.
\newblock Spatialreasoner: Towards explicit and generalizable 3d spatial
  reasoning.
\newblock \emph{Advances in Neural Information Processing Systems},
  38:\penalty0 140751--140774, 2026{\natexlab{b}}.

\bibitem[Ning et~al.(2024)Ning, Peng, Wang, Sun, Liu, Yuille, Kortylewski, and
  Wang]{ning2024part}
C.~Ning, J.~Peng, J.~Wang, Y.~Sun, Y.~Liu, A.~Yuille, A.~Kortylewski, and
  A.~Wang.
\newblock Part321: Recognizing 3d object parts from a 2d image using 1-shot
  annotations, 2024.
\newblock URL \url{https://openreview.net/forum?id=jdFoxDnBwY}.

\bibitem[Oquab et~al.(2023)Oquab, Darcet, Moutakanni, Vo, Szafraniec, Khalidov,
  Fernandez, Haziza, Massa, El-Nouby, et~al.]{oquab2023dinov2}
M.~Oquab, T.~Darcet, T.~Moutakanni, H.~Vo, M.~Szafraniec, V.~Khalidov,
  P.~Fernandez, D.~Haziza, F.~Massa, A.~El-Nouby, et~al.
\newblock Dinov2: Learning robust visual features without supervision.
\newblock \emph{arXiv preprint arXiv:2304.07193}, 2023.

\bibitem[Puig et~al.(2023)Puig, Undersander, Szot, Cote, Partsey, Yang, Desai,
  Clegg, Hlavac, Min, Gervet, Vondrus, Berges, Turner, Maksymets, Kira,
  Kalakrishnan, Malik, Chaplot, Jain, Batra, Rai, and
  Mottaghi]{puig2023habitat3}
X.~Puig, E.~Undersander, A.~Szot, M.~D. Cote, R.~Partsey, J.~Yang, R.~Desai,
  A.~W. Clegg, M.~Hlavac, T.~Min, T.~Gervet, V.~Vondrus, V.-P. Berges,
  J.~Turner, O.~Maksymets, Z.~Kira, M.~Kalakrishnan, J.~Malik, D.~S. Chaplot,
  U.~Jain, D.~Batra, A.~Rai, and R.~Mottaghi.
\newblock Habitat 3.0: A co-habitat for humans, avatars and robots, 2023.

\bibitem[Qiu et~al.(2017{\natexlab{a}})Qiu, Zhong, Zhang, Qiao, Xiao, Kim, and
  Wang]{qiu2017unrealcv}
W.~Qiu, F.~Zhong, Y.~Zhang, S.~Qiao, Z.~Xiao, T.~S. Kim, and Y.~Wang.
\newblock Unrealcv: Virtual worlds for computer vision.
\newblock In \emph{Proceedings of the 25th ACM international conference on
  Multimedia}, pages 1221--1224, 2017{\natexlab{a}}.

\bibitem[Qiu et~al.(2017{\natexlab{b}})Qiu, Zhong, Zhang, Qiao, Xiao, Kim, and
  Wang]{unrealcv}
W.~Qiu, F.~Zhong, Y.~Zhang, S.~Qiao, Z.~Xiao, T.~S. Kim, and Y.~Wang.
\newblock Unrealcv: Virtual worlds for computer vision.
\newblock In \emph{Proceedings of the 25th ACM international conference on
  Multimedia}, pages 1221--1224, 2017{\natexlab{b}}.

\bibitem[Radford et~al.(2021)Radford, Kim, Hallacy, Ramesh, Goh, Agarwal,
  Sastry, Askell, Mishkin, Clark, et~al.]{radford2021learning}
A.~Radford, J.~W. Kim, C.~Hallacy, A.~Ramesh, G.~Goh, S.~Agarwal, G.~Sastry,
  A.~Askell, P.~Mishkin, J.~Clark, et~al.
\newblock Learning transferable visual models from natural language
  supervision.
\newblock In \emph{International conference on machine learning}, pages
  8748--8763. PmLR, 2021.

\bibitem[Raistrick et~al.(2023)Raistrick, Lipson, Ma, Mei, Wang, Zuo, Kayan,
  Wen, Han, Wang, Newell, Law, Goyal, Yang, and Deng]{infinigen-natural}
A.~Raistrick, L.~Lipson, Z.~Ma, L.~Mei, M.~Wang, Y.~Zuo, K.~Kayan, H.~Wen,
  B.~Han, Y.~Wang, A.~Newell, H.~Law, A.~Goyal, K.~Yang, and J.~Deng.
\newblock Infinite photorealistic worlds using procedural generation.
\newblock In \emph{Proceedings of the IEEE/CVF Conference on Computer Vision
  and Pattern Recognition}, pages 12630--12641, 2023.

\bibitem[Raistrick et~al.(2024)Raistrick, Mei, Kayan, Yan, Zuo, Han, Wen,
  Parakh, Alexandropoulos, Lipson, Ma, and Deng]{infinigen-indoor}
A.~Raistrick, L.~Mei, K.~Kayan, D.~Yan, Y.~Zuo, B.~Han, H.~Wen, M.~Parakh,
  S.~Alexandropoulos, L.~Lipson, Z.~Ma, and J.~Deng.
\newblock Infinigen indoors: Photorealistic indoor scenes using procedural
  generation.
\newblock In \emph{Proceedings of the IEEE/CVF Conference on Computer Vision
  and Pattern Recognition (CVPR)}, pages 21783--21794, June 2024.

\bibitem[Ray et~al.(2024)Ray, Duan, Brown, Tan, Bashkirova, Hendrix, Ehsani,
  Kembhavi, Plummer, Krishna, et~al.]{ray2024sat}
A.~Ray, J.~Duan, E.~Brown, R.~Tan, D.~Bashkirova, R.~Hendrix, K.~Ehsani,
  A.~Kembhavi, B.~A. Plummer, R.~Krishna, et~al.
\newblock Sat: Dynamic spatial aptitude training for multimodal language
  models.
\newblock \emph{arXiv preprint arXiv:2412.07755}, 2024.

\bibitem[Roberts et~al.(2021)Roberts, Ramapuram, Ranjan, Kumar, Bautista,
  Paczan, Webb, and Susskind]{roberts2021hypersim}
M.~Roberts, J.~Ramapuram, A.~Ranjan, A.~Kumar, M.~A. Bautista, N.~Paczan,
  R.~Webb, and J.~M. Susskind.
\newblock Hypersim: A photorealistic synthetic dataset for holistic indoor
  scene understanding.
\newblock In \emph{Proceedings of the IEEE/CVF international conference on
  computer vision}, pages 10912--10922, 2021.

\bibitem[Ros et~al.(2016)Ros, Sellart, Materzynska, Vazquez, and
  Lopez]{ros2016synthia}
G.~Ros, L.~Sellart, J.~Materzynska, D.~Vazquez, and A.~M. Lopez.
\newblock The synthia dataset: A large collection of synthetic images for
  semantic segmentation of urban scenes.
\newblock In \emph{Proceedings of the IEEE conference on computer vision and
  pattern recognition}, pages 3234--3243, 2016.

\bibitem[Ruiz et~al.(2022)Ruiz, Kortylewski, Qiu, Xie, Bargal, Yuille, and
  Sclaroff]{Ruiz_2022_CVPR}
N.~Ruiz, A.~Kortylewski, W.~Qiu, C.~Xie, S.~A. Bargal, A.~Yuille, and
  S.~Sclaroff.
\newblock Simulated adversarial testing of face recognition models.
\newblock In \emph{Proceedings of the IEEE/CVF Conference on Computer Vision
  and Pattern Recognition (CVPR)}, pages 4145--4155, June 2022.

\bibitem[Shu et~al.(2020)Shu, Liu, Qiu, and Yuille]{shu2020identifying}
M.~Shu, C.~Liu, W.~Qiu, and A.~Yuille.
\newblock Identifying model weakness with adversarial examiner.
\newblock In \emph{Proceedings of the AAAI conference on artificial
  intelligence}, volume~34, pages 11998--12006, 2020.

\bibitem[Sim{\'e}oni et~al.(2025)Sim{\'e}oni, Vo, Seitzer, Baldassarre, Oquab,
  Jose, Khalidov, Szafraniec, Yi, Ramamonjisoa, et~al.]{simeoni2025dinov3}
O.~Sim{\'e}oni, H.~V. Vo, M.~Seitzer, F.~Baldassarre, M.~Oquab, C.~Jose,
  V.~Khalidov, M.~Szafraniec, S.~Yi, M.~Ramamonjisoa, et~al.
\newblock Dinov3.
\newblock \emph{arXiv preprint arXiv:2508.10104}, 2025.

\bibitem[Slim et~al.(2025)Slim, Li, Li, Ahmed, Ayman, Upadhyay, Abdelreheem,
  Prajapati, Pothigara, Wonka, et~al.]{slim2023_3dcompatplus}
H.~Slim, X.~Li, Y.~Li, M.~Ahmed, M.~Ayman, U.~Upadhyay, A.~Abdelreheem,
  A.~Prajapati, S.~Pothigara, P.~Wonka, et~al.
\newblock 3dcompat++: An improved large-scale 3d vision dataset for
  compositional recognition.
\newblock \emph{IEEE Transactions on Pattern Analysis and Machine
  Intelligence}, 2025.

\bibitem[Sun et~al.(2025)Sun, Liu, Gu, Lim, Bhat, Tombari, Li, Haber, and
  Wu]{sun2025layoutvlm}
F.-Y. Sun, W.~Liu, S.~Gu, D.~Lim, G.~Bhat, F.~Tombari, M.~Li, N.~Haber, and
  J.~Wu.
\newblock Layoutvlm: Differentiable optimization of 3d layout via
  vision-language models.
\newblock In \emph{Proceedings of the Computer Vision and Pattern Recognition
  Conference}, pages 29469--29478, 2025.

\bibitem[Todorov et~al.(2012)Todorov, Erez, and Tassa]{todorov2012mujoco}
E.~Todorov, T.~Erez, and Y.~Tassa.
\newblock Mujoco: A physics engine for model-based control.
\newblock In \emph{2012 IEEE/RSJ international conference on intelligent robots
  and systems}, pages 5026--5033. IEEE, 2012.

\bibitem[Tosi et~al.(2021)Tosi, Liao, Schmitt, and Geiger]{Tosi2021CVPR}
F.~Tosi, Y.~Liao, C.~Schmitt, and A.~Geiger.
\newblock Smd-nets: Stereo mixture density networks.
\newblock In \emph{Conference on Computer Vision and Pattern Recognition
  (CVPR)}, 2021.

\bibitem[Wang et~al.(2025{\natexlab{a}})Wang, Xue, and
  Gao]{wang2025infinibench}
H.~Wang, Q.~Xue, and W.~Gao.
\newblock Infinibench: Infinite benchmarking for visual spatial reasoning with
  customizable scene complexity.
\newblock \emph{arXiv preprint arXiv:2511.18200}, 2025{\natexlab{a}}.

\bibitem[Wang et~al.(2025{\natexlab{b}})Wang, Chen, Karaev, Vedaldi, Rupprecht,
  and Novotny]{wang2025vggt}
J.~Wang, M.~Chen, N.~Karaev, A.~Vedaldi, C.~Rupprecht, and D.~Novotny.
\newblock Vggt: Visual geometry grounded transformer.
\newblock In \emph{Proceedings of the Computer Vision and Pattern Recognition
  Conference}, pages 5294--5306, 2025{\natexlab{b}}.

\bibitem[Wang et~al.(2024)Wang, Leroy, Cabon, Chidlovskii, and
  Revaud]{wang2024dust3r}
S.~Wang, V.~Leroy, Y.~Cabon, B.~Chidlovskii, and J.~Revaud.
\newblock Dust3r: Geometric 3d vision made easy.
\newblock In \emph{Proceedings of the IEEE/CVF conference on computer vision
  and pattern recognition}, pages 20697--20709, 2024.

\bibitem[Wang et~al.(2023)Wang, Ma, Li, Kortylewski, and Yuille]{wang20233d}
X.~Wang, W.~Ma, Z.~Li, A.~Kortylewski, and A.~L. Yuille.
\newblock 3d-aware visual question answering about parts, poses and occlusions.
\newblock \emph{Advances in Neural Information Processing Systems},
  36:\penalty0 58717--58735, 2023.

\bibitem[Wang et~al.(2025{\natexlab{c}})Wang, Ma, Zhang, de~Melo, Chen, and
  Yuille]{wang2025spatial457}
X.~Wang, W.~Ma, T.~Zhang, C.~M. de~Melo, J.~Chen, and A.~Yuille.
\newblock Spatial457: A diagnostic benchmark for 6d spatial reasoning of large
  mutimodal models.
\newblock In \emph{Proceedings of the Computer Vision and Pattern Recognition
  Conference}, pages 24669--24679, 2025{\natexlab{c}}.

\bibitem[Yang et~al.(2024)Yang, Sun, Weihs, VanderBilt, Herrasti, Han, Wu,
  Haber, Krishna, Liu, et~al.]{yang2024holodeck}
Y.~Yang, F.-Y. Sun, L.~Weihs, E.~VanderBilt, A.~Herrasti, W.~Han, J.~Wu,
  N.~Haber, R.~Krishna, L.~Liu, et~al.
\newblock Holodeck: Language guided generation of 3d embodied ai environments.
\newblock In \emph{Proceedings of the IEEE/CVF Conference on Computer Vision
  and Pattern Recognition}, pages 16227--16237, 2024.

\bibitem[Ye et~al.()Ye, Ren, Zhuang, He, Liang, Yang, Dogra, Zhong, Liu,
  Benavente, et~al.]{yesimworld}
X.~Ye, J.~Ren, Y.~Zhuang, X.~He, Y.~Liang, Y.~Yang, M.~Dogra, X.~Zhong, E.~Liu,
  K.~Benavente, et~al.
\newblock Simworld: An open-ended simulator for agents in physical and social
  worlds.
\newblock In \emph{The Thirty-ninth Annual Conference on Neural Information
  Processing Systems}.

\bibitem[Yi et~al.(2019)Yi, Gan, Li, Kohli, Wu, Torralba, and
  Tenenbaum]{yi2019clevrer}
K.~Yi, C.~Gan, Y.~Li, P.~Kohli, J.~Wu, A.~Torralba, and J.~B. Tenenbaum.
\newblock Clevrer: Collision events for video representation and reasoning.
\newblock \emph{arXiv preprint arXiv:1910.01442}, 2019.

\bibitem[Yin et~al.(2026)Yin, Ge, Wang, Li, Black, Darrell, Kanazawa, and
  Feng]{yin2026visionasinversegraphicsagentinterleavedmultimodal}
S.~Yin, J.~Ge, Z.~Z. Wang, X.~Li, M.~J. Black, T.~Darrell, A.~Kanazawa, and
  H.~Feng.
\newblock Vision-as-inverse-graphics agent via interleaved multimodal
  reasoning, 2026.
\newblock URL \url{https://arxiv.org/abs/2601.11109}.

\bibitem[Zhang et~al.(2025)Zhang, Liu, Li, Wen, Guan, Wang, and
  Nie]{zhang2025spatial}
H.~Zhang, M.~Liu, Z.~Li, H.~Wen, W.~Guan, Y.~Wang, and L.~Nie.
\newblock Spatial understanding from videos: Structured prompts meet simulation
  data.
\newblock \emph{arXiv preprint arXiv:2506.03642}, 2025.

\bibitem[Zhang et~al.(2024)Zhang, Herrmann, Hur, Jampani, Darrell, Cole, Sun,
  and Yang]{zhang2024monst3r}
J.~Zhang, C.~Herrmann, J.~Hur, V.~Jampani, T.~Darrell, F.~Cole, D.~Sun, and
  M.-H. Yang.
\newblock Monst3r: A simple approach for estimating geometry in the presence of
  motion.
\newblock \emph{arXiv preprint arXiv:2410.03825}, 2024.

\bibitem[Zhao et~al.(2022)Zhao, Yu, Ma, Yu, Mei, Wang, He, Yuille, and
  Kortylewski]{zhao2022ood}
B.~Zhao, S.~Yu, W.~Ma, M.~Yu, S.~Mei, A.~Wang, J.~He, A.~Yuille, and
  A.~Kortylewski.
\newblock Ood-cv: A benchmark for robustness to out-of-distribution shifts of
  individual nuisances in natural images.
\newblock In \emph{European conference on computer vision}, pages 163--180.
  Springer, 2022.

\bibitem[Zhao et~al.(2024)Zhao, Wang, Ma, Jesslen, Yang, Yu, Zendel, Theobalt,
  Yuille, and Kortylewski]{zhao2024ood}
B.~Zhao, J.~Wang, W.~Ma, A.~Jesslen, S.~Yang, S.~Yu, O.~Zendel, C.~Theobalt,
  A.~L. Yuille, and A.~Kortylewski.
\newblock Ood-cv-v2: An extended benchmark for robustness to
  out-of-distribution shifts of individual nuisances in natural images.
\newblock \emph{IEEE Transactions on Pattern Analysis and Machine
  Intelligence}, 46\penalty0 (12):\penalty0 11104--11118, 2024.

\bibitem[Zhong et~al.(2025)Zhong, Wu, Wang, Chen, Ci, Li, and
  Wang]{zhong2025unrealzoo}
F.~Zhong, K.~Wu, C.~Wang, H.~Chen, H.~Ci, Z.~Li, and Y.~Wang.
\newblock Unrealzoo: Enriching photo-realistic virtual worlds for embodied ai.
\newblock In \emph{Proceedings of the IEEE/CVF International Conference on
  Computer Vision}, pages 5769--5779, 2025.

\bibitem[Zhou et~al.(2021)Zhou, Wei, Wang, Shen, Xie, Yuille, and
  Kong]{zhou2021ibot}
J.~Zhou, C.~Wei, H.~Wang, W.~Shen, C.~Xie, A.~Yuille, and T.~Kong.
\newblock ibot: Image bert pre-training with online tokenizer.
\newblock \emph{arXiv preprint arXiv:2111.07832}, 2021.

\end{thebibliography}

\clearpage

\appendix
\onecolumn
\section*{Appendix}

\begin{comment}
\noindent
\begin{minipage}[t]{0.48\textwidth}
    \begin{itemize}[label={},leftmargin=0pt,itemsep=1.2ex]
        \item a
        \item b
        \item c
    \end{itemize}
\end{minipage}
\hfill
\begin{minipage}[t]{0.48\textwidth}
    \begin{itemize}[label={},leftmargin=0pt,itemsep=1.2ex]
        \item Section~\ref{sec:supp_annotations}: data annotations and release plan
        \item b
        \item c
    \end{itemize}
\end{minipage}
\end{comment}

\section{Technical Details} \label{sec:technical}

\subsection{Comprehensive 2D and 3D Ground Truths} \label{sec:supp_gt}

\begin{figure}[t]
    \centering
    \includegraphics[width=\textwidth]{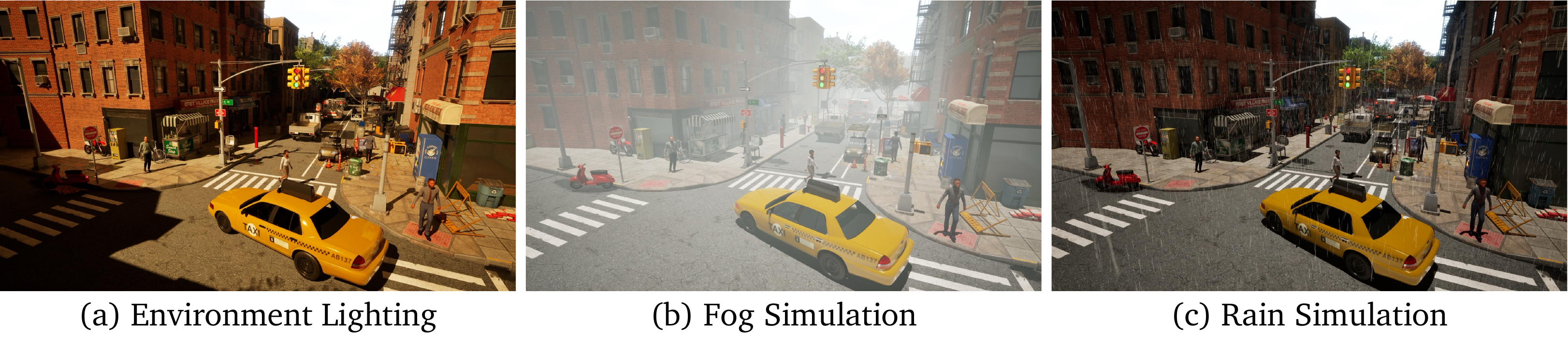}
    \includegraphics[width=\textwidth]{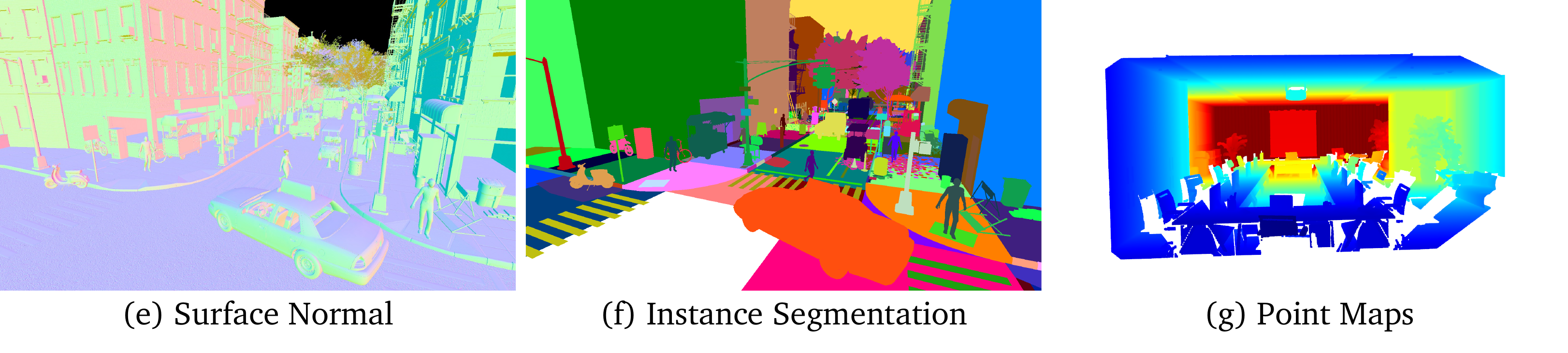}
    \includegraphics[width=0.866\textwidth]{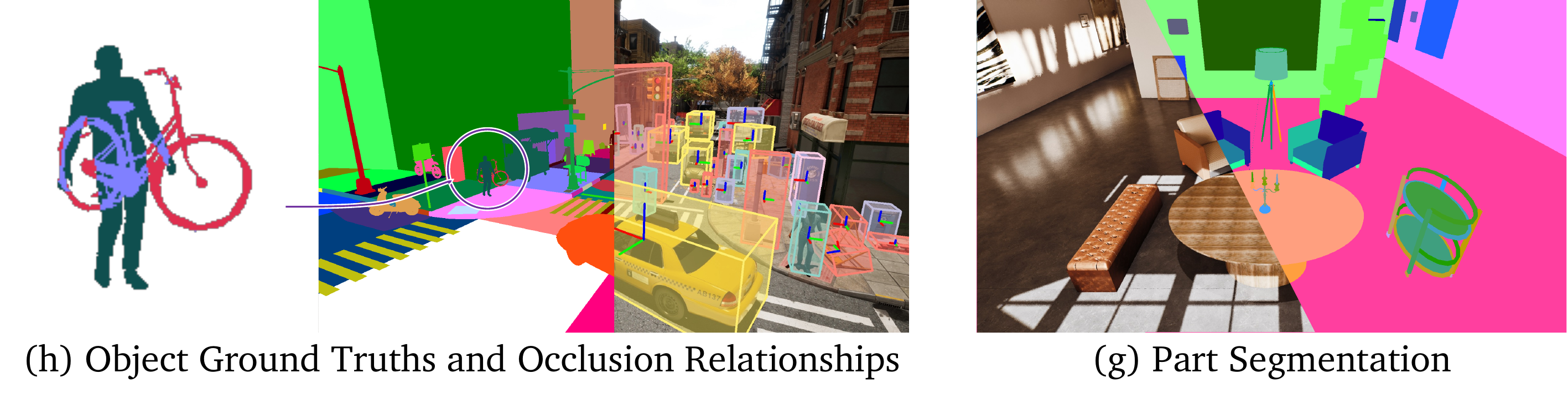}
    \caption{\textbf{Rich 2D and 3D ground truths in LychSim.}}
    \label{fig:supp_gt}
\end{figure}

LychSim extracts comprehensive ground-truth annotations at the scene, view, and object levels. This hierarchical structure facilitates rigorous model training and evaluation by capturing a full spectrum of visual factors, ranging from global environmental conditions to fine-grained, occluded or part-level object geometries. Please refer to Figure~\ref{fig:supp_gt} for visualizations of various 2D and 3D ground truths.

\paragraph{Scene-level ground truths.} To capture global environmental variations, LychSim records macroscopic parameters, including: (1) directional and ambient lighting configurations; (2) quantitative fog visibility metrics; and (3) rain simulation parameters. These variables enable systematic model evaluation under challenging weather and illumination shifts.

\paragraph{View-level ground truths.} At the camera level, LychSim renders pixel-aligned spatial annotations, encompassing: (1) depth maps; (2) instance segmentation masks; (3) surface normals; and (4) rendered dense point maps that output per-pixel 3D vertex coordinates to natively support modern 3D learning pipelines.

\paragraph{Object-level ground truths.} For fine-grained analysis, LychSim extracts object-centric annotations: (1) 2D and 3D bounding boxes; (2) semantically aligned 3D poses; (3) occlusion and truncation ratios computed via 3D geometric projections to quantify visibility beyond the image plane; and (4) part segmentations. The engine supports both general mesh parts and semantic part segmentations for assets with defined hierarchies, such as in DST-3DPart~\cite{ning2024part} or 3DCoMPaT++~\cite{slim2023_3dcompatplus}.

\subsection{Model Context Protocol (MCP)} \label{sec:supp_mcp}

We build the Model Context Protocol (MCP) on top of our LychSim Python API (see Figure~\ref{fig:interactive_flow}), establishing a seamless bridge between agentic LLMs and high-fidelity, interactive 3D simulation. With MCP integraion, LychSim empowers LLMs to autonomously execute tools that query spatial states, place objects, and capture real-time visual feedback.
%
% This closed-loop design is critical for advanced spatial reasoning, as it allows agents to visually verify their actions and self-correct physical implausibilities.
%
Furthermore, to support the demanding, iterative nature of agentic planning, we enable the parallel rendering of multiple camera viewpoints. This optimization effectively reduces communication overhead, allowing models to rapidly assess the scene from various angles without bottlenecking the simulation.

\paragraph{Tool schema.}
Our native MCP server implementation is built upon FastMCP, utilizing a standard JSON schema for tool definitions. However, in practice, XML-based schemas are often favored for LLM tool calling, as XML provides explicit start and end boundaries that significantly reduce syntax formatting errors (e.g., missing brackets or unescaped quotes) during generation.
To bridge this gap, XML interactions is supported through third-party MCP clients. In widely used environments such as Claude Code and Cursor, the native LychSim JSON schema is automatically translated into an XML format for the LLM. Subsequently, the LLM-generated XML tool calls are processed by lenient parsers, translated back into standard JSON payloads, and securely routed back to the LychSim engine for execution. See example MCP JSON schema in Code~\ref{lst:schema}.

% \subsection{Scene Setup} \label{sec:technical_scene}

\subsection{Procedural Rules and Object Annotations} \label{sec:supp_annotations}

\paragraph{Interactive annotation tool.} We introduce a dedicated annotation toolset within LychSim to enable the creation of procedural rules directly inside the Unreal Editor. The interactive interface allows users to define spatial relationships and semantic regions simply by clicking to select starting and ending anchor objects. With a single command, these custom annotations are recorded and exported for use in the data generation pipeline. The tool natively supports three distinct geometric data types to cover various procedural rules: (1) directed straight lines, (2) spline curves for smooth trajectories, and (3) square areas for regional zoning. An screenshot of this interface is shown in Figure~\ref{fig:supp_procedural}.

\paragraph{Procedural rules.} We categorize our procedural annotations into four distinct types: (1) road and street areas (outdoor regions), (2) person-navigable areas (applicable indoors and outdoors), (3) vehicle trajectories (outdoor pathways), and (4) pedestrian trajectories (outdoor pathways). These four primitives provide the essential spatial priors needed to generate highly diverse, physically plausible layouts across varying domains.
Specifically, outdoor environments are heavily governed by structural semantics; vehicles often adhere to road networks, and crowds naturally follow sidewalks or crosswalks. By combining region and trajectory rules, we ensure that outdoor generation respects these semantic contextual boundaries. Conversely, indoor environments typically feature less rigid, open-ended movement spaces. We primarily rely on broader navigable areas to flexibly guide indoor layout generation, such as furniture placement or agent navigation.
Lastly these foundational rules serve as the foundation for controllable data synthesis. By systematically perturbing or overloading these spatial constraints, we easily generate complex out-of-distribution (OOD) scenarios, such as extreme dense scenes, heavy occlusion, or uncommon camera viewpoints, to rigorously evaluate model robustness.

\paragraph{Object annotations.} To ensure consistency across the diverse asset library, we annotate each 3D object with the following attributes: (1) a semantic category label; (2) a canonical object scale, ensuring uniform sizing across disparate mesh sources; (3) a standardized sampling offset (set to the bottom-center of the mesh) to guarantee consistent ground-level alignment during procedural placement; (4) a canonical 3D pose alignment, ensuring that the forward-facing vectors of all objects are identically oriented; and (5) an LLM-generated, descriptive text caption, such as "a red tractor with a rust-streaked engine cover" or "a brown teddy bear with button eyes." Collectively, these annotations are essential for extracting semantically aligned 3D ground-truth poses and enabling precise, language-driven object placement during automated scene generation.

% \subsection{Interactive Planning} \label{sec:interactive_planning}

\section{Code and Data Releases}

We are committed to making the LychSim framework fully accessible to the broader vision community.

\paragraph{Code and data access.} We will publicly release the complete C++ and Python source code, the MCP server implementation, the list of 3D assets, and all associated procedural rules and object-level pose alignments. We will also host comprehensive documentation, API references, and quick-start tutorials to ensure a seamless onboarding experience.

\paragraph{License.} The LychSim source code will be open-sourced under MIT license to encourage broader academic and industry adoption. All newly curated data annotations will be released under Creative Commons Attribution 4.0 International license (CC BY 4.0), while the underlying 3D meshes will remain subject to the standard Fab Asset Marketplace terms.

\paragraph{Project maintenance and community.} To ensure the long-term viability of the project, our core development team is dedicated to an active maintenance plan. We will continuously monitor the public repository to address bug reports, ensure compatibility with future engine updates, and actively review community pull requests. We welcome community feedback and collaborative contributions to expand LychSim's functionalities and environmental diversity over time.

\section{Supplementary Documents}

\begin{figure}[t]
    \centering
    \includegraphics[width=\linewidth]{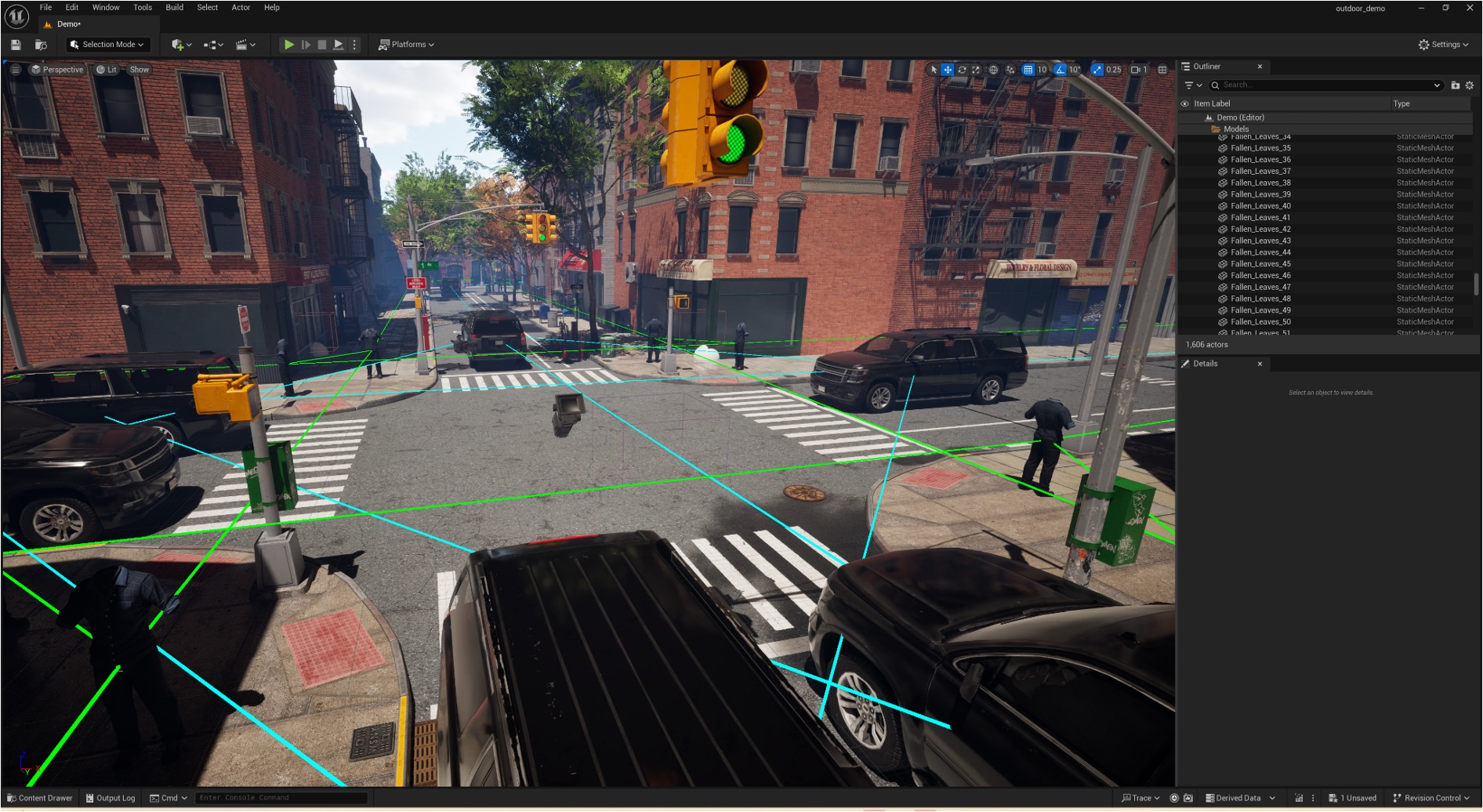}
    \caption{\textbf{Interactive annotation tool and procedural rules in LychSim.}}
    \label{fig:supp_procedural}
\end{figure}

\begin{enumerate}
\item Interactive annotation tool and procedural rules in LychSim: Figure~\ref{fig:supp_procedural}.
\item Example MCP tool schema: Code~\ref{lst:schema}.
\item Claude skill for scene planning: Code~\ref{lst:plan_room}.
\item User input for loft office specification: Code~\ref{lst:office}.
\end{enumerate}

\clearpage

\clearpage

\begin{lightcode}[title=Example MCP tool schema,label={lst:schema}]{json}
{
  "name": "spawn_object",
  "description": "Spawn a new object into the LychSim scene.\n\nReturns:\n    `{\"status\": \"ok\"}` on success, or\n    `{\"status\": \"<error_code>\"}` describing what went wrong\n    (e.g. `\"object_with_same_name_already_exists\"`,\n    `\"failed_to_spawn_actor\"`, `\"unknown_argument_format\"`).",
  "inputSchema": {
    "type": "object",
    "properties": {
      "obj_id": {
        "type": "string",
        "title": "Obj Id",
        "description": "A unique name for the new object (e.g. `\"Table_1\"`). Must not collide with an existing object ID in the scene — call `list_objects` first to check."
      },
      "obj_path": {
        "type": "string",
        "title": "Obj Path",
        "description": "Unreal Engine asset path for the mesh or blueprint to spawn (e.g. `\"/Game/Assets/Mesh/SM_Table\"`). Use `list_objects` + `get_object_location` to inspect what is already in the scene, but asset paths come from the project's Content directory, not from scene queries."
      },
      "location": {
        "type": "array",
        "items": {
          "type": "number"
        },
        "title": "Location",
        "description": "World-space `[x, y, z]` in centimeters (left-handed, Z-up). Defaults to the world origin `[0, 0, 0]`.",
        "default": [0.0, 0.0, 0.0]
      },
      "rotation": {
        "type": "array",
        "items": {
          "type": "number"
        },
        "title": "Rotation",
        "description": "`[pitch, yaw, roll]` in degrees. Defaults to `[0, 0, 0]`.",
        "default": [0.0, 0.0, 0.0]
      },
      "scale": {
        "type": "number",
        "title": "Scale",
        "description": "Uniform scale factor. Defaults to `1.0`.",
        "default": 1.0
      },
      "collision_handling": {
        "type": "string",
        "title": "Collision Handling",
        "description": "How to handle spawn-time collisions. `\"default\"` — always spawn; `\"skip_if_colliding\"` — do not spawn if the location overlaps existing geometry; `\"adjust_if_possible\"` — try to nudge the object to a free spot, but fail if none is found.",
        "default": "default"
      },
      "lock_rotation": {
        "type": "boolean",
        "title": "Lock Rotation",
        "description": "If `true`, lock the actor's rotation after spawning (useful for static props).",
        "default": false
      }
    },
    "required": [
      "obj_id",
      "obj_path"
    ]
  }
}
\end{lightcode}

\clearpage

\begin{lightcode}[title=Claude skill for scene planning: SKILL.md,label={lst:plan_room}]{markdown}
---
name: plan_scene
description: Plan and place furniture in a LychSim Unreal Engine scene from a high-level scene spec (assets, room geometry, layout requirements). Iterates by querying the scene, placing objects, and verifying visually with camera screenshots.
---

# Plan Scene

Use this skill when the user asks to "lay out", "furnish", "plan", "set up", or "decorate" a LychSim scene from a markdown spec (e.g. `office.md`). The goal is to take an asset list + room geometry + layout requirements and produce a coherent, visually verified scene.

## Available MCP tools

You will use the `mcp__lychsim__*` tools:

- **State queries**: `list_objects`, `get_object_location`, `get_object_rotation`, `get_camera_location`, `get_camera_rotation`
- **Spawning / editing**: `add_object`, `set_object_location`, `set_object_rotation`, `update_object`, `delete_object`
- **Sizing**: `get_mesh_extent` (may fail on some meshes — see Pitfalls)
- **Camera**: `set_camera_location`, `set_camera_rotation`, `get_camera_lit`

## Workflow

1. **Read the spec.** Parse asset paths, room geometry (floor corners, X/Y/Z ranges), layout requirements, placement options.

2. **Snapshot the current state.** In parallel: `list_objects`, `get_camera_location`, `get_camera_rotation`, then `get_camera_lit`. The room is rarely empty — there are usually persistent scene props you should not delete.

3. **Plan zones, not coordinates.** Sketch the layout in zones (desk area, reading nook, plant corners) before computing positions. Functional groupings beat scattered placement.

4. **Place anchors first.** Spawn the largest anchoring objects (table, soft chair) before stacking smaller items on/around them. Use `collision_handling: "adjust_if_possible"` from the spec.

5. **Stack using estimated heights.** A standard desk top is at floor Z + ~75cm. If `get_mesh_extent` works, use it; otherwise estimate. Place monitor/books/vase Y between desk Y±40 so they land on the desk, not a neighboring chair.

6. **Place chairs with rotation last.** Don't trust your first guess at chair facing — see Mesh Forward Direction below.

7. **Verify from multiple angles.** Top-down (`pitch=-89`, high Z) for layout. Side views for chair orientations. Wide-angle from a corner for the final beauty shot.

8. **Iterate.** Fix overlaps, wrong-facing chairs, items inside furniture. The user expects you to look at every screenshot critically and self-correct.

9. **Restore the final camera pose.** When the scene is done, move the camera to the **"Final camera location and rotation"** values specified in the spec (e.g. `office.md`). This is the canonical hero-shot pose the user expects to see when they next open the scene. Use `set_camera_location` and `set_camera_rotation`, then take one last `get_camera_lit` to confirm.

10. **Desktop items.** Place desktop items at the table-top Z, not on the floor.

## Coordinate system

- Centimeters, **left-handed, Z-up**.
- `yaw=0` → forward = **+X**. `yaw=90` → +Y. `yaw=-90` → -Y. `yaw=180` → -X.
- Floor is typically at Z = -20 in LoftOffice scenes; furniture spawn locations sit at floor Z (objects pivot from their base for most static meshes, but not all — verify).

## Placement options (from spec)

Map the spec's options to tool args:

| Spec key             | How to apply                                                 |
|----------------------|--------------------------------------------------------------|
| `collision_handling` | Pass directly to `add_object` (`adjust_if_possible` is safest) |
| `skip_existing`      | Call `list_objects` first; skip ids already in scene         |
| `clear_scene_first`  | If true, delete only the objects YOU previously spawned, not the existing scene props |
| `lock_rotation`      | Pass to `add_object` as `lock_rotation`                      |

The spec may also include a **"Final camera location and rotation"** section. Always restore the camera to this pose as the very last step (see step 9 in the workflow).

## Camera tips

- **Top-down overview**: camera at scene-center XY, Z \approx 250, rotation `[-89, 0, 0]`. Best for verifying layout and chair facing.
- **Wide corner shot**: camera at one floor corner, Z \approx 250, pitch \approx -25, yaw pointing diagonally across the room. Best for the final hero image.
- **Side view of a single object**: camera offset 100–200cm from the object along a perpendicular axis, Z \approx 80, slight downward pitch.
- Always restore or note the camera pose so the user can return to it.

## Output

After placement, present a brief summary:

- Bullet list of placed objects (id, asset, location)
- One or two camera screenshots from a wide angle
- Any issues you noticed and self-corrected
- Suggestions for the user (lighting, additional props, alternative arrangements)
\end{lightcode}

\clearpage

\begin{lightcode}[title=User input for loft office specification: loft\_office.md,label={lst:office}]{markdown}
# Scene: Loft Office

## Assets

- Chair: /Game/LoftOffice/Meshes/SM_Chair_3.SM_Chair_3
- Office chair: /Game/LoftOffice/Meshes/SM_Chair_1.SM_Chair_1
- Soft chair: /Game/LoftOffice/Meshes/SM_Armchair_2.SM_Armchair_2
- Vase:/Game/LoftOffice/Meshes/SM_Vase_1.SM_Vase_1
- Table: /Game/LoftOffice/Meshes/SM_Table_2.SM_Table_2
- Monitor: /Game/LoftOffice/Meshes/SM_Monitor_2.SM_Monitor_2
- Plant: /Game/LoftOffice/Meshes/SM_Plant_3.SM_Plant_3
- Floor lamp: /Game/LoftOffice/Meshes/SM_Floor_Lamp.SM_Floor_Lamp
- Stack of books: /Game/LoftOffice/Meshes/SM_Stack_of_Books_2.SM_Stack_of_Books_2
- Document: /Game/LoftOffice/Meshes/SM_Document_case.SM_Document_case
- Small cabinet: /Game/LoftOffice/Meshes/SM_Drawer.SM_Drawer

## Room Geometry

Floor corners (X Y Z, centimeters, Z-up):
```
420 -410 -20
420  180 -20
869  180 -20
869 -410 -20
```

Floor Z: -20
X range: 420 -- 869 (depth ~449 cm)
Y range: -410 -- 180 (width ~590 cm)

Final camera location: 260 -300 165
Final camera rotation: 0 -13 24

## Layout Requirements

- 1 table with 1 monitor and 1 stack of books on top
- 3 chairs around the table (mix of office chair and regular chair)
- 1 soft chair in a corner as a reading nook
- 2 plants (scale ~5.0 to make them floor-sized)
- 1 floor lamp next to the soft chair
- 1 vase on the table

## Placement Options

- collision_handling: adjust_if_possible
- skip_existing: true
- clear_scene_first: false
- lock_rotation: false

## Style Notes

- Arrange furniture in functional groupings (desk area, reading corner), not scattered randomly.
- Face chairs toward their associated table or focal point.
- Keep walkways clear (~80 cm minimum between furniture groups).
- Place plants near walls or corners, not in the middle of the room.
- Stack objects (monitor, books, vase on table) using mesh extents to compute correct Z offsets.
\end{lightcode}

\end{document}